\documentclass[runningheads]{llncs}

\usepackage{eccv}

\usepackage{eccvabbrv}
\usepackage[section]{placeins}
\usepackage{microtype}
\microtypesetup{expansion=false}
\usepackage{graphicx}
\usepackage{subcaption}
\usepackage{multirow}

\usepackage{booktabs} 

\usepackage{amsmath}
\usepackage{amssymb}
\usepackage{mathtools}
\usepackage{bbm}

\usepackage[textsize=tiny]{todonotes}

\usepackage[accsupp]{axessibility}  

\usepackage{hyperref}
\usepackage[capitalize,noabbrev]{cleveref}

\spnewtheorem{assumption}[theorem]{Assumption}{\bfseries}{\rmfamily}

\newcommand{\ours}{\textbf{$\text{M}^3\text{Bench}$\xspace}}
\usepackage{orcidlink}
\begin{document}

\title{Evaluating and Understanding Model Editing for Medical Vision Language Models}

\titlerunning{Evaluating and Understanding Model Editing for Medical VLMs}

\author{Guli Zhu\textsuperscript{*}\inst{1} \and
Chenwei Wu\textsuperscript{*}\inst{1} \and
Liyue Shen\inst{1}}

\authorrunning{G. Zhu and C. Wu et al.}

\institute{EECS, University of Michigan, Ann Arbor, MI 48105, USA \\
\email{\{gulizhu, chenweiw, liyues\}@umich.edu}}

\maketitle

\begingroup
\renewcommand{\thefootnote}{*}
\footnotetext{Equal contribution, order determined by coin flip.}
\endgroup

\begin{abstract}
Model editing promises a fast, targeted way to correct post-deployment mistakes in medical vision--language models (VLMs) without costly retraining. However, existing multimodal model editing benchmarks focus on general-purpose tasks and do not reflect realistic clinical domain requirements and variability. To address this, we introduce \ours, a clinically grounded benchmark for multimodal model editing that evaluates whether an edit remains reliable, precise and generalizable under the challenges of image and text variation, modality and protocol shifts, clinical knowledge composition, and temporal progression. \ours~contains 16,276 questions spanning diverse anatomy, modalities, and specialties, and supports both single and sequential edits. 
By evaluating 4 representative editors across 6 medical and general VLMs, we indicate that no method excels across all criteria. Gradient-based editors achieve strong transfer but suffer from catastrophic locality violations, whereas memory-based methods preserve locality but lack compositional generality and exhibit high backbone-dependent hyperparameter sensitivity. We further attribute these failures to the latent space geometry of VLMs and how different editing methods shift its landscape. Overall, \ours~establishes a rigorous clinical stress test for multimodal model editing and offers actionable guidance for safer post-deployment adaptation. The benchmark is publicly available at \url{https://github.com/BioMed-AI-Lab-U-Michgan/M3Bench}.

\end{abstract}

\section{Introduction}
\label{sec:intro}

\begin{figure}[t]
  \centering
  \begin{subfigure}[t]{0.65\linewidth}
    \centering
    \includegraphics[width=\linewidth]{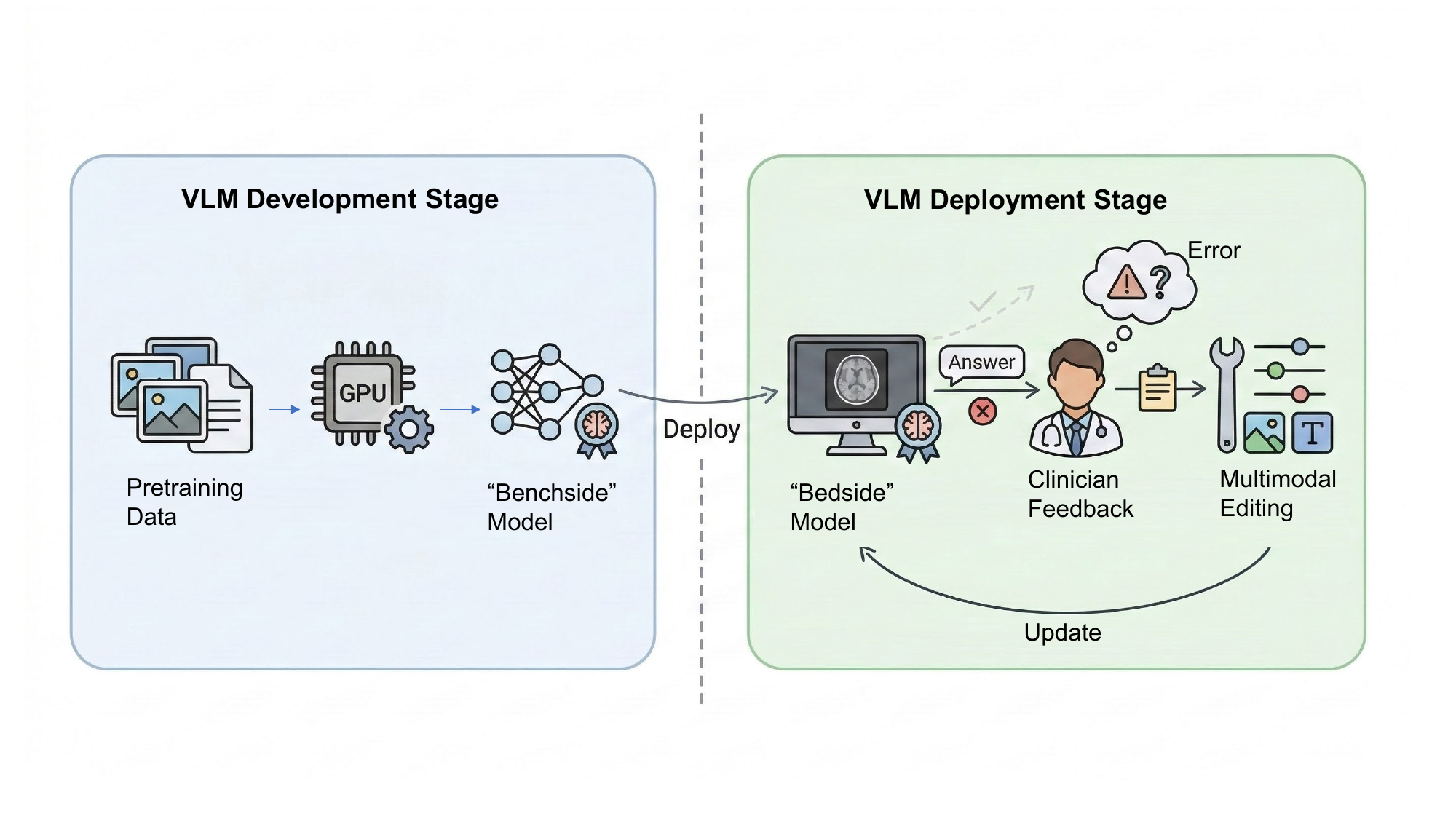}
    \caption{} 
    \label{fig:intro_workflow_a}
  \end{subfigure}
  \hfill
  \begin{subfigure}[t]{0.34\linewidth}
    \centering
    \includegraphics[width=\linewidth]{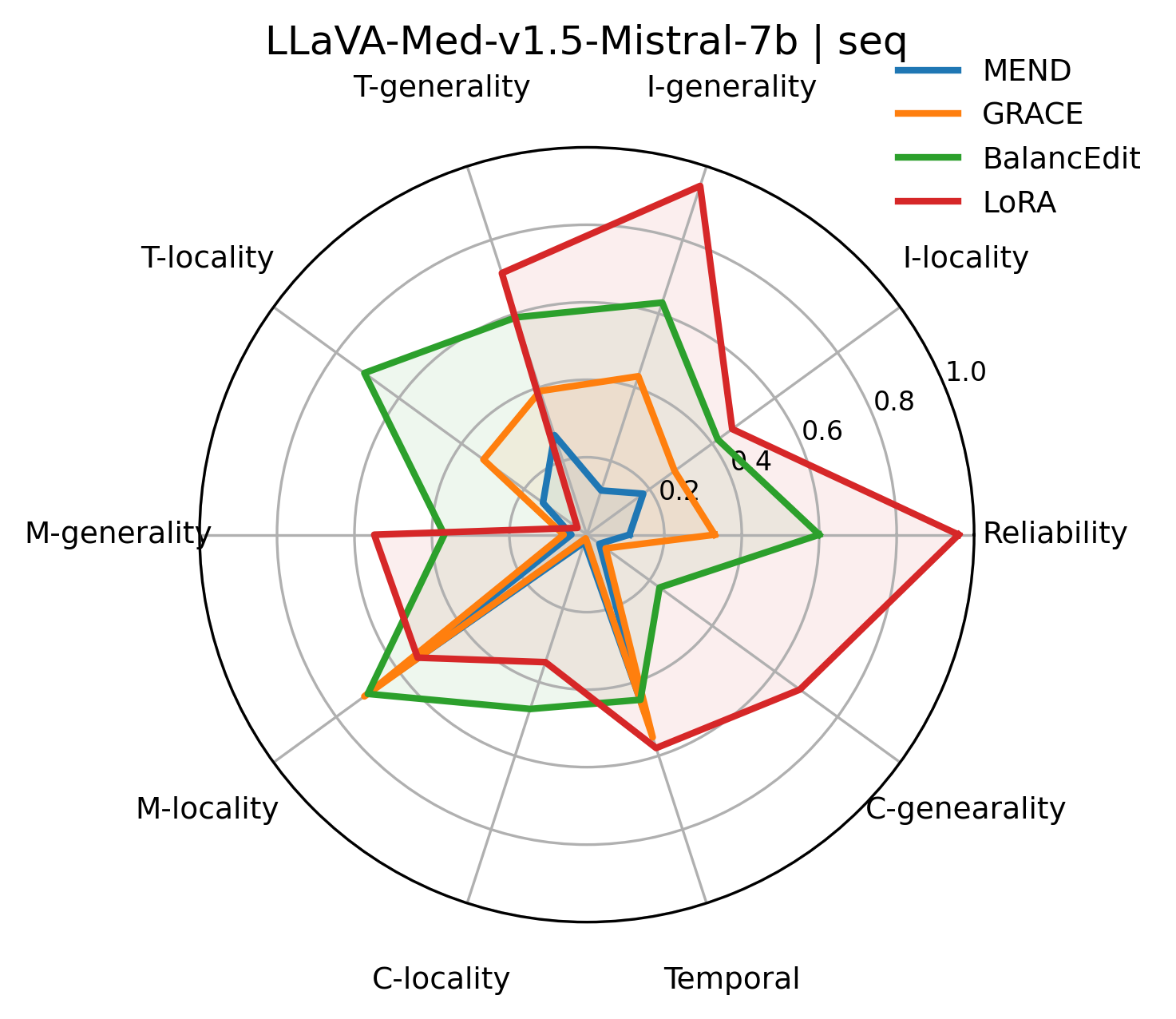}
    \caption{} 
    \label{fig:intro_workflow_b}
  \end{subfigure}

  \caption{\textbf{Post-deployment workflow and model editing performance.}
  \textbf{(a)} \textbf{Post-deployment workflow of multimodal model editing} with clinician-in-the-loop feedback.
  \textbf{Left:} model development stage with extensive model training.
  \textbf{Right:} model deployment stage with lightweight and targeted model editing to correct errors on the fly.
  \textbf{(b)} Editing performance radar (sequential-edit) summarizing performance across our \ours\ evaluation dimensions for LLaVA-Med-v1.5-Mistral-7b.}
  \vspace{-18pt}
  \label{fig:intro_workflow_radar}
\end{figure}

Vision--Language Models (VLMs) hold great promise in supporting multimodal clinical workflows such as automated radiology report generation and real-time surgical assistance~\cite{acosta2022multimodal,cheng2026surgxbench}. However, as shown in Fig.~\ref{fig:intro_workflow_a}, real-world deployment of these large-scale models is not a one-time milestone: once a model enters clinical practice, it is exposed to continuously changing patient data and clinical protocols that may not be observed during model development. 
As a result, even well-trained models will inevitably make errors after deployment, potentially leading to serious consequences, such as failing to recognize rare diseases, misinterpreting complex patient cases involving compositional conditions, or incorrectly analyzing studies acquired with new scanners~\cite{khera2023automation,alsaad2024multimodal,liu2020reporting,yu2022external}.

Addressing such post-deployment errors through global model updates, such as full fine-tuning or retraining, is often impractical, as they are computationally expensive, data-intensive, time-consuming, and may degrade prior model capabilities~\cite{rosenthal2025rethinking}.
\emph{Model Editing} has therefore emerged as a promising alternative for targeted intervention~\cite{wang2024knowledge}, enabling precise corrections by updating a small subset of model parameters or incorporating external parameter memory, while keeping the rest of the model unchanged. 
While these techniques have been increasingly studied for general-domain LLMs and VLMs~\cite{du2025mmke}, \textbf{how model editing behaves under realistic multimodal clinical tasks} remains largely underexplored. 
Editing methods that ``succeed'' under existing general-domain evaluation may fail catastrophically in real-world clinical applications. 
Most existing benchmarks evaluate three axes: \emph{Reliability} (whether edits correct targeted errors), \emph{Locality} (whether edits avoid unintended changes), and \emph{Generality} (whether edits transfer to related cases)~\cite{meng2022rome,memla2024mkeb}. 
However, these assessments rely on simplified tasks that underestimate the complexity and variability of real-world clinical settings. 
For example, when evaluating ``generality'', simplified tasks of image perturbation through noise addition do not adequately reflect the practical clinical relevance, such as similar pathological findings observed across patients or different views of the same patient's X-Ray scans.
These limitations call for a reformulation of evaluation tasks and metrics to enable clinically grounded assessment.

In this study, we introduce a new \emph{\textbf{M}ultimodal \textbf{M}edical \textbf{M}odel-editing Benchmark} (\ours) designed to holistically evaluate and understand multimodal model editing for LLM-based vision–language models in clinical settings. 
As shown in Fig.~\ref{fig:model}, \ours~is built around the core editing objectives of Reliability, Locality, and Generality, but redefines them through clinically meaningful tasks. 
In particular, we evaluate whether an edit remains precise and generalizable under four clinically motivated settings: 
(1) \emph{textual variation},
(2) \emph{visual variation},
(3) \emph{modality and protocol variation},
and (4) \emph{clinical composition}.
Additionally, we assess \emph{temporal consistency}, measuring whether model knowledge remains coherent across longitudinal prior–current comparisons where disease progression is clinically relevant.

Through evaluations on 10 different tasks across 4 popular medical VLM (LLaVA-Med~\cite{li2023llava}, Huatuo7B, Huatuo34B~\cite{chen2024huatuogptvisioninjectingmedicalvisual}, BioMed-Qwen~\cite{cheng2025domainadaptiveposttrainingmultimodallarge}) and 2 general VLM backbones (Qwen3.5-2B~\cite{team2026qwen3}, Janus-Pro-7B~\cite{chen2025janus}), we holistically conduct the first systematic study to benchmark existing model editing methods, which we taxonomize into \textit{memory-based approaches} (e.g., GRACE~\cite{hartvigsen2023aginggracelifelongmodel}, BalancEdit~\cite{guo2025balanceditdynamicallybalancinggeneralitylocality}) and \textit{gradient-based approaches} (e.g., LoRA~\cite{hu2022lora}, MEND~\cite{mitchell2022fastmodeleditingscale}), under clinical deployment challenges. 
As shown in Fig.~\ref{fig:intro_workflow_b}, our evaluation shows that no single editing method consistently outperforms others across all tasks.
Instead, distinct tradeoffs emerge between editing families. 
Specifically, gradient-based editors often achieve strong reliability and successfully transfer knowledge to semantically similar cases, but suffer from exceptionally poor locality. 
Conversely, memory-based methods like BalancEdit excel as the most well-rounded, though less generalizable than gradient-based methods, and are highly sensitive to hyperparameters relying on backbone models. 
Furthermore, across all tested methods, temporal consistency and clinical composition consistently prove to be amongst the most challenging dimensions for current editors to resolve.

Moreover, to understand the mechanisms underlying these discrepancies, we analyze how editing methods reshape the internal geometry of medical concepts in the VLM latent space. 
We find that pre-trained medical VLM representations are highly concentrated in a narrow cone, making concept separation and localized precise editing intrinsically difficult. 
This geometric bottleneck drives the observed behavioral differences: gradient-based editors tend to induce global concept drifts in the latent space, whereas memory-based editors apply edits only within a tightly gated activation region, which preserves locality but can miss interleaved concepts and limits generalization in realistic settings.
Besides, the degree of cone tightness varies significantly across VLM backbones, which explains the observed hyperparameter sensitivity in memory-based methods. 
Overall, our investigation provides insights and actionable guidance on when and how to apply different editing methods to enhance trustworthy clinical deployment.
Our key contributions are three-fold:
\begin{itemize}
    \item \textbf{New benchmark} that provides comprehensive evaluation for multimodal model editing methods with clinically grounded evaluation dimensions by defining 10 criteria across 6 representative medical and general VLM models.
    \item \textbf{New findings} from our systematic evaluation and taxonomy of editing families, including tradeoffs between gradient-based and memory-based editing approaches and disclosing previously underexplored failure modes.
    \item \textbf{New insights} from our mechanistic analyses linking editing behavior to representation geometry and sensitivity, yielding actionable guidance for developing effective and reliable multimodal model editing in clinical workflow.
\end{itemize}

\section{Preliminaries}
\label{sec:related}

\subsection{Vision-Language Models for Medical Applications}
Medical Vision-Language Models (VLMs) demonstrate strong potential for multimodal clinical tasks, including disease diagnosis, radiology report generation, and image-grounded question answering~\cite{liu2023medical}. Most existing VLMs adopt an encoder-projector-LLM architecture, where visual features extracted from a vision encoder (e.g., CLIP~\cite{radford2021learning}) are mapped into the language embedding space via a lightweight projector. These visual tokens are then interleaved with textual embeddings, allowing the LLM to generate clinical responses autoregressively in a sequence. 
These models are typically adapted from general-purpose VLMs to the medical domain via medical instruction tuning~\cite{cheng2025domainadaptiveposttrainingmultimodallarge}. 
While proficient in medical terminology, these models remain prone to hallucinations and are highly sensitive to data distribution shifts~\cite{liu2023medical,finlayson2021clinician}. To ensure a comprehensive evaluation, we evaluate our \ours~across a diverse suite of 4 representative medical VLMs: LLaVA-Med~\cite{li2023llava}, HuatuoGPT-Vision (7B/34B)~\cite{chen2024huatuogptvisioninjectingmedicalvisual}, and BioMed-Qwen~\cite{cheng2025domainadaptiveposttrainingmultimodallarge} and 2 general VLM backbones (Qwen3.5-2B~\cite{team2026qwen3}, Janus-Pro-7B~\cite{chen2025janus}). This selection spans varying model architectures and scales, pretraining data composition, and alignment recipes.

\subsection{Knowledge Editing}
Knowledge editing, also known as model editing, aims to efficiently update a pretrained model to produce a desired response for a targeted fact without compromising its broader capabilities~\cite{meng2022rome}. 
More concretely, given an edit request $(x_e, y_e)$, an editing algorithm $\mathcal{A}$ is supposed to update the original model, denoted as $f_{\theta}$ with $\theta$ indicating model parameters, to obtain an updated model $f_{\theta'}$ with new parameters $\theta'$ that precisely satisfies this edit request using minimal supervision and compute, without retraining the full model. 
Here we categorize four representative model editing methods for our evaluation:

\paragraph{Gradient-based Editing.}
\textbf{MEND}, introduced for LLM editing, learns a meta-network that predicts a weight update from the gradient of a single edit example, avoiding iterative per-edit optimization and reducing latency~\cite{mitchell2022fastmodeleditingscale}. \textbf{LoRA} is a general approach for parameter-efficient model update by learning low-rank update matrices inserted into the selected layers.
Thus, in the model editing task, we also consider LoRA as a feasible approach enabling lightweight edits and easier storage of multiple updates~\cite{hu2022lora}.

\paragraph{Memory-based Editing.}
These approaches avoid directly rewriting backbone model weights by caching parameters of a chosen layer aside and selectively activating them. During training, \textbf{Grace} stores modified layer weights for different edit inputs as value-keys pairs in a codebook; if a new key falls within the radius of a previously stored key, then the two keys are merged and the new edit is finetuned sequentially on the previously stored weights~\cite{hartvigsen2023aginggracelifelongmodel}. During inference, it treats the hidden state of a test question as a query; the edited patch is triggered only if the query falls within a preset radius of a stored key. If the query is outside this radius, the model defers to its original, unedited path. 
\textbf{BalancEdit} follows this same retrieve-and-apply principle but makes the radius dynamic across samples~\cite{guo2025balanceditdynamicallybalancinggeneralitylocality}. To manage the trade-off between generality and locality, it learns a boundary using two anchor samples: a positive anchor (text rephrasings of the edit), and a negative anchor (a null/black image). A hyperparameter $\alpha$ controls this ``trigger radius'' by calculating a weighted sum of $1-\alpha$ times the distance from query to positive anchor, and $\alpha$ times the distance to negative anchor.
\vspace{-10pt}
\subsection{Editing Evaluation}
Evaluating model editing for medical-domain VLMs introduces unique challenges, distinct from the general-domain LLMs editing evaluation~\cite{meng2022rome,memla2024mkeb}, 
First, in multimodal settings, editing can fail in ways that do not arise in text-only models: an edit may become only text-dependent rather than image-grounded, which purely ignores visual cues, or induce inconsistent behavior across images that share similar clinical concepts but with patient or modality variation.
Therefore, it is highly crucial that we systematically stress-test common factors such as whether edited knowledge generalizes across same-concept images, or whether edits affect unrelated concepts present within the same image. Second, high-stake medical applications introduce more complexity and variability yet low tolerance for harmful model behavior than the general-domain counterparts~\cite{du2025mmke}. For example, simple paraphrases to measure text generality will fail as the same medical text could appear in different synonyms, abbreviations, or report-style phrasing. Going beyond concurrent works~\cite{memla2024mkeb, medmkeb2025}, we introduce a clinically grounded evaluation suite that directly accounts for such real-world deployment requirements of medical multimodal assistants.
\vspace{-10pt}

\section{Benchmark}
\label{sec:setup}

We study post-deployment multimodal model editing for medical-domain VLMs used in medical question answering. 
Suppose a deployed model $f_{\theta}$ maps an image $I$ and a query $q$ to generate an answer $\hat{y}=f_{\theta}(I,q)$. 
When a failure or mistaken output is identified in practice, clinicians issue an edit request $e=(I,q,y^{\star})$, where $y^{\star}$ is the desired corrected answer for the target input $(I,q)$. Given an editing algorithm $\mathcal{A}$, we obtain an edited model 
$f_{\theta'}=\mathcal{A}(f_{\theta};e)$. A clinically useful edit should satisfy three requirements: 
~\textbf{Reliability} enforces the correction on the target instance, i.e., $f_{\theta'}(I,q)=y^{\star}$;
~\textbf{Locality} requires that the edit does not corrupt already-correct model knowledge and behaviors; and
~\textbf{Generality} requires the corrected knowledge can transfer to clinically equivalent variants in other cases.

\label{sec:benchmark}
\begin{figure}[t]
  \centering
  \includegraphics[width=\textwidth]{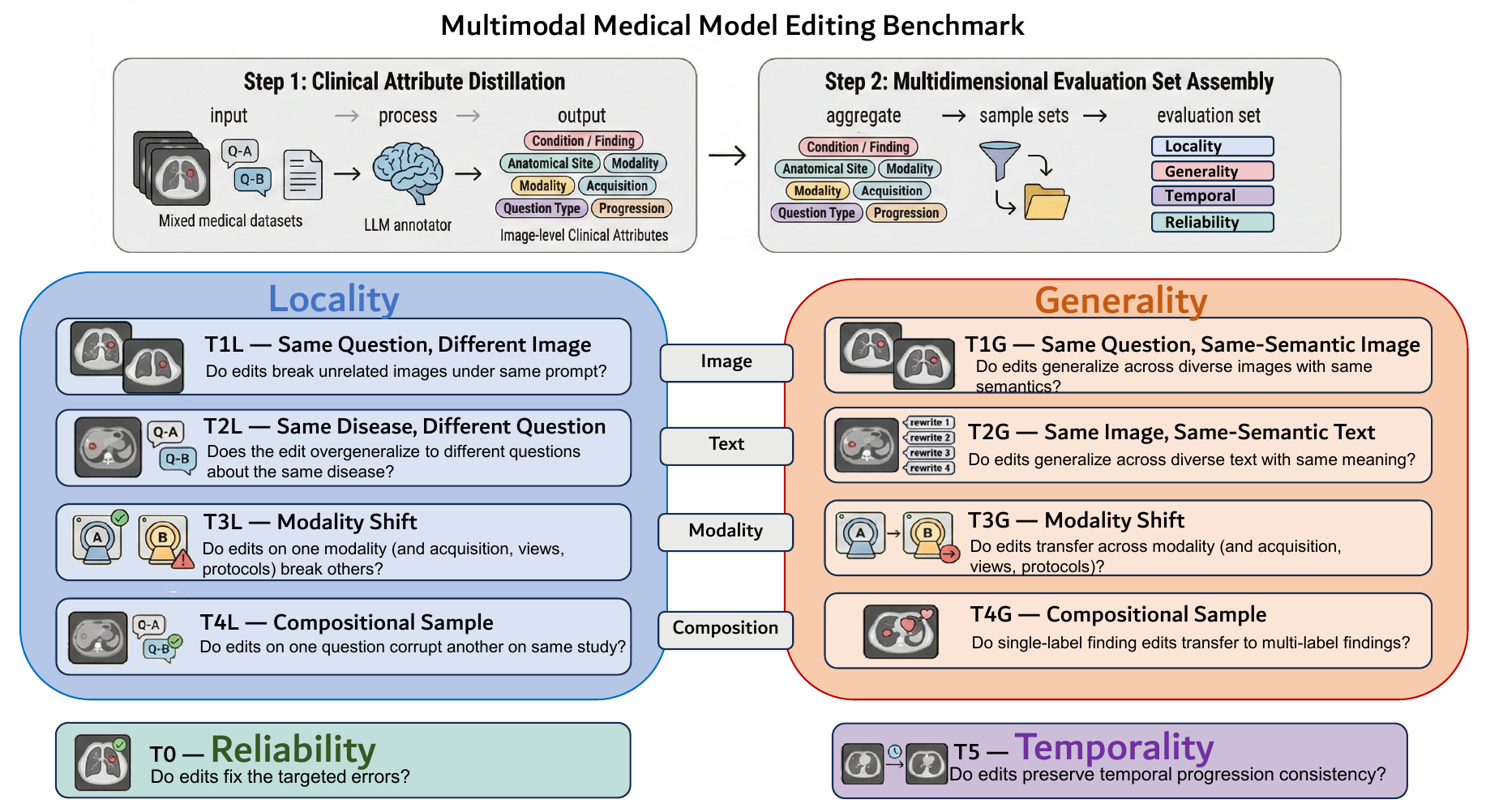}
  \vspace{-18pt}
  \caption{\textbf{Construction pipeline of our proposed Multimodal Medical Model Editing Benchmark (\ours)}. We define 10 clinically grounded evaluation tasks spanning multiple dimensions of reliability, locality, generality, and temporality.}
  \label{fig:model}
  \vspace{-14pt}
\end{figure}

\subsection{Data Sources and Construction}
 \label{sec:data-construct}
 We start building our new benchmark (\ours) by leveraging various existing medical VQA and image-text dataset resources, including VQA-RAD, PMC-VQA, PadChest-GR, and SLAKE~\cite{lau2018dataset,zhang2023pmc,de2025padchest,liu2021slake}.
Our construction pipeline follows a two-stage process, as illustrated in Fig.~\ref{fig:model}:

\noindent \textbf{Clinical Attribute Distillation.} We employ LLMs as expert annotators to process raw QA pairs or associated clinical notes. For every image-text pair, the LLM extracts a standardized attribute list including medical conditions, anatomical sites, modalities, acquisition context, question types, and progression status. 
This structured representation serves as a common “schema” that aligns examples across datasets and reduces annotation noise from wording and formatting differences.

\noindent \textbf{Multi-dimensional Evaluation Set Assembly.} After aggregating all attributes for the same image, we can programmatically construct evaluation sets $\mathcal{D}$ by isolating specific clinical variables. 
For example, we select samples that are clinically similar but differ along a specific axis (e.g., a new patient image with the same finding; the same patient finding under a different imaging view or protocol).
The resulting benchmark contains 16,276 questions covering 9 anatomical sites and 4 imaging modalities.

\subsection{Task Suite}
 \label{sec:task-suite}
 Our key motivation is to move beyond general purpose evaluation tasks to reflect the critical and clinically meaningful challenges that the models would face after real-world clinical deployment. 
 As shown in Fig.~\ref{fig:model}, we expand common definitions of locality and generality along 4 clinically motivated axes of variation. Reliability and a novel metric of temporal consistency are also included, totaling 10 tasks.
 
\paragraph{Axis 0:}\textbf{T0 -- Reliability.}
We first verify whether the edit successfully corrects the originally observed failure on the target input $(I,q)$.
\paragraph{Axis 1: Image.}
 \textbf{T1L -- Image Locality (same question, different semantic-unrelated images).} 
 Clinicians often reuse templated questions across many patients. A dangerous failure mode in VLMs~\cite{hu2025enhancing} occurs when edits overfit the text and ignore visual evidence. We therefore keep the question fixed and evaluate on unrelated images where the correct answer should remain unchanged. 
\textbf{T1G -- Image Generality (same question, same-semantic images).}
 Conversely, the same findings or disease may recur across different patients. An edit that only repairs the single edited image is of limited clinical value. We therefore evaluate the same question on new images that share the same underlying visual concepts (e.g., exact matches on distilled attributes such as finding and anatomical site, image with perturbations). This measures whether the edit transfers to semantically equivalent patient cases.
\paragraph{Axis 2: Text.}
 \textbf{T2L -- Text Locality (same image, different clinical-attribute question).}
 Real users often ask multiple questions about the same study. An edit should not corrupt other correct answers within same image context. We therefore evaluate on other questions about $I$, which target different clinical attributes compared with edit question $q$.
\textbf{T2G -- Text Generality (same image, semantic-preserving rewrites).}
 Medical language varies widely: abbreviations, shorthand, formatting differences, negation patterns, and different reasoning styles can all express the same meaning. We test whether an edit remains valid under multiple semantically equivalent rewrites of $q$ (e.g., paraphrase; abbreviation/synonym; rephrased question form; telegraphic note style; mix-in irrelevant but factual clinical text). 

\paragraph{Axis 3: Modality.}
\textbf{T3L -- Modality Locality.}
The same patient finding can appear different under changes in scanner, protocol, view, or modality. We apply the edit under images from an acquisition setting $A$ and evaluate the same question under images obtained from a shifted setting $B$ while keeping other attributes fixed. Locality tests whether the edit avoids regressing on cases under $B$ that were already correct before editing. \textbf{T3G -- Modality Generality.}
Under the same $A\rightarrow B$ shift, generality tests whether the correction transfers to concept-matched cases under $B$ that were previously answered incorrectly. 

\paragraph{Axis 4: Clinical composition (multi-label findings).}
 \textbf{T4G -- Compositional Generality.}
 Clinical cases are rarely “single-concept”: multiple findings often co-occur in one patient study. We therefore test whether an edit learned from a single-finding context transfers when the target finding appears alongside additional findings. This captures compositional transfer rather than isolated correction.
\textbf{T4L -- Compositional Locality.}
 On multi-finding images, an edit should not introduce contradictions with other correct knowledge about the same study. We thus evaluate questions about other findings on the same study image that were previously answered correctly, probing whether the edit preserves internal consistency under composition.
 
\paragraph{Axis 5:}\textbf{T5 -- Temporal consistency.}
Clinical interpretation must remain coherent across follow-ups, where progression or regression matters. We therefore edit on an earlier timepoint and evaluate on the corresponding follow-up study for the same patient. This task tests that edits do not introduce temporal contradictions in longitudinal progression.

\paragraph{Metrics.} We score reliability as the post edit accuracy; We score locality 1 - the percentage a model edit breaks previously non-target correct answers; We score generality as the percentage an edit successfully corrects previously wrong answers; we score temporality as  1 - percentage an edit at earlier time introduces factual errors in later follow-up studies. We also compute an overall harmonic mean across 10 tasks, penalizing methods with extreme failures on any dimensions. For all metrics, higher values indicate better performance.

\section{Experiments}
\label{sec:exp}
We conducted main single and sequential knowledge editing experiments on 4 medical VLMs with different parameter sizes and architectures, including LLaVA-Med~\cite{li2023llava}, HuatuoGPT-Vision (7B/34B)~\cite{chen2024huatuogptvisioninjectingmedicalvisual}, and BioMed-Qwen~\cite{cheng2025domainadaptiveposttrainingmultimodallarge}. We additionally include 1 general domain VLM Qwen3.5-2B~\cite{team2026qwen3} and 1 general domain native multimodal model Janus Pro-7B~\cite{chen2025janus} results in Appendix. As detailed in Sec.~\ref{sec:related}, we selected 4 representative knowledge editing methods and taxonomize into 2 groups, gradient based methods (MEND~\cite{mitchell2022fastmodeleditingscale}, Lora~\cite{hu2022lora}) and memory-based methods (Grace~\cite{hartvigsen2023aginggracelifelongmodel}, BalancEdit~\cite{guo2025balanceditdynamicallybalancinggeneralitylocality}). All experiments are performed on one NVIDIA A100 GPU with 80GB VRAM. The batch size of all experiments is uniformly set to 1. All method-specific hyperparameters are tuned to ensure optimal performance. Notably, our evaluation method was implemented from scratch to avoid inheriting issues like teacher forcing from existing editing libraries. All of our model results are free autoregressive generation. Tab.~\ref{tab:seqresults} is also consistent with~\cite{yang2025mirage} that real-world editing performance is substantially lower than teacher-forced eval scores. Other details are included in the Appendix. 

\begin{figure}[t]
  \centering
  \includegraphics[width=\linewidth]{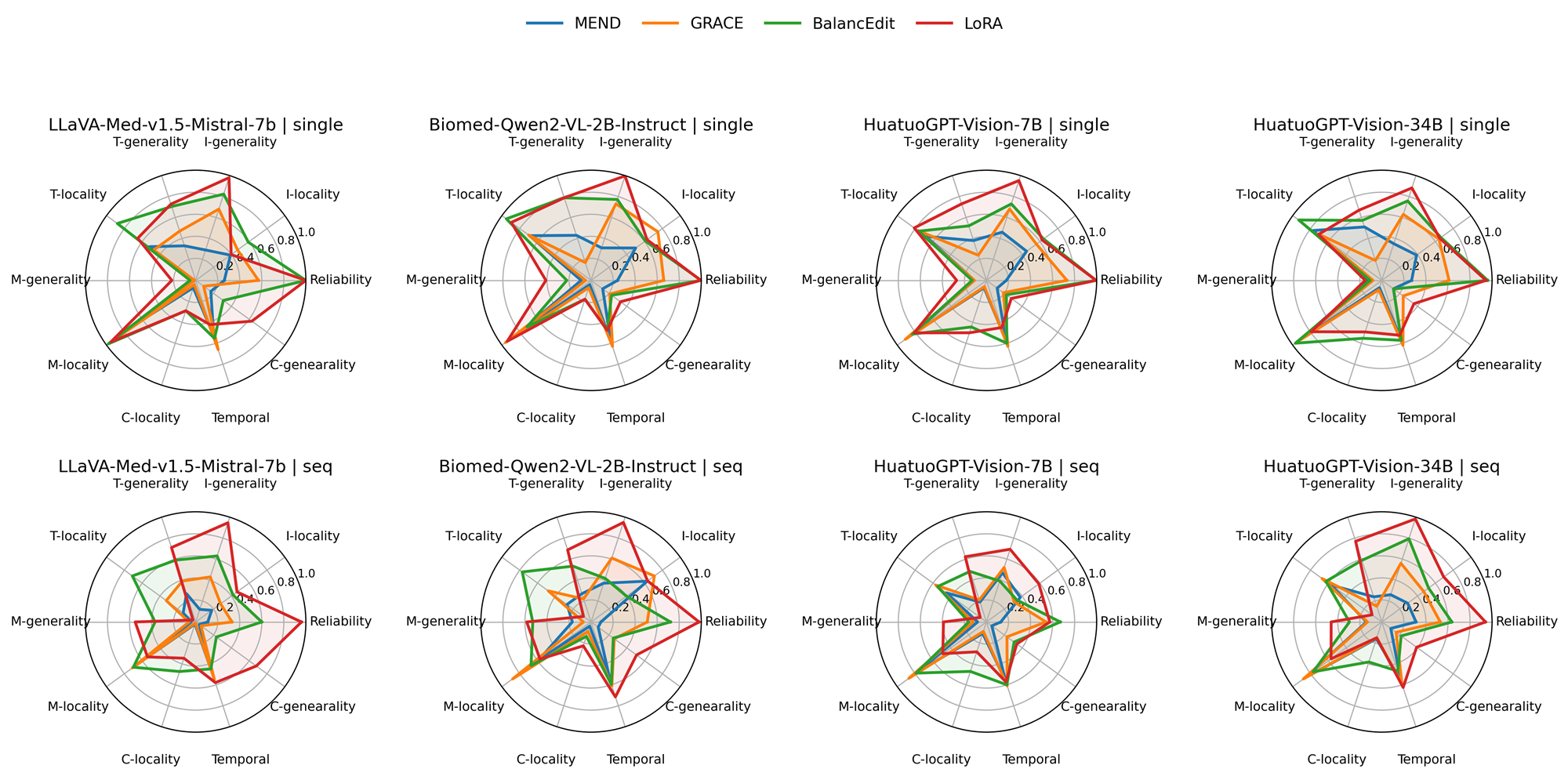}

  \caption{Radar plot summarizing \textbf{single and sequential editing performance} across 10 tasks and 4 medical VLM backbones. General VLM results in Appendix.}
  \label{fig:radar_results}
  
\end{figure}

\subsection{Main Results}
\label{sec:main-results}
As summarized in Table \ref{tab:seqresults} and Figure \ref{fig:radar_results}, our systematic evaluation across four medical VLMs reveals distinct performance profiles among the tested editing methods. Notably, across all backbones, we observe that \textbf{no single editing paradigm dominates every criterion.} 
Distinct trade-offs emerge between editing families, as discussed below.

\textbf{LoRA attains excellent reliability and generality, but locality collapses.} Gradient-based editing via LoRA delivers the strongest reliability and typically high generality scores (e.g., I-generality and T-generality in Tab.~\ref{tab:seqresults}). However, it frequently exhibits exceptionally poor locality under clinically plausible perturbations (notably T-locality), indicating concerning unintended behavior changes after editing. 

\textbf{BalancEdit stands out as the most well-rounded, but composition, temporality, and overall generality remain challenging.} BalancEdit (BE) achieves the strongest overall performance on most backbones (Tab.~\ref{tab:seqresults}, Overall), reflecting consistently competitive reliability and locality. However, clinical composition and temporality remain challenging for all editors: BE improves over alternatives on several composition-related locality scores (e.g., C-locality) yet absolute compositional transfer remains low (e.g., C-generality), and temporal metrics are far from solved. In addition, during experimentation, we found memory-based method performance to be highly sensitive to hyperparameter choices (the edit decision radius discussed in Sec. \ref{sec:related}) and the optimal values differ in magnitudes across different backbones, making it hard for practical generalization.

\section{Analysis}
 To understand why different editing families exhibit sharply different tradeoffs in M$^3$Bench, we study editing through a geometric lens: how edits reshape the latent organization of medical concepts, and how this interacts with model architecture and editing hyperparameters.

\begin{table}[t]
\centering
\caption{\textbf{Sequential editing performance} across models and methods (higher is better). Values rounded to 2 decimals.In bold means best performing method.}
\label{tab:seqresults}
\resizebox{\columnwidth}{!}{%
\begin{tabular}{lcccccccccccccccc}
\toprule
& \multicolumn{4}{c}{\textbf{LLaVA-Med}} & \multicolumn{4}{c}{\textbf{BioQwen}} & \multicolumn{4}{c}{\textbf{Huatuo-7b}} & \multicolumn{4}{c}{\textbf{Huatuo-34b}} \\
\cmidrule(lr){2-5}\cmidrule(lr){6-9}\cmidrule(lr){10-13}\cmidrule(lr){14-17}
\textbf{Metric}
& MEND & GRACE & BE & LORA
& MEND & GRACE & BE & LORA
& MEND & GRACE & BE & LORA
& MEND & GRACE & BE & LORA \\
\midrule
\textbf{Reliability}   & 0.11 & 0.33 & 0.60 & 0.96 & 0.09 & 0.51 & 0.72 & 0.98 & 0.13 & 0.54 & 0.67 & 0.58 & 0.31 & 0.53 & 0.64 & 0.94 \\
\hline
\textbf{I-locality}    & 0.18 & 0.28 & 0.42 & 0.46 & 0.63 & 0.71 & 0.40 & 0.63 & 0.38 & 0.31 & 0.34 & 0.59 & 0.29 & 0.47 & 0.52 & 0.69 \\
\textbf{I-generality}  & 0.12 & 0.43 & 0.63 & 0.95 & 0.37 & 0.61 & 0.42 & 0.95 & 0.47 & 0.52 & 0.39 & 0.69 & 0.26 & 0.56 & 0.79 & 0.98 \\
\hline
\textbf{T-generality}  & 0.27 & 0.39 & 0.59 & 0.71 & 0.27 & 0.22 & 0.53 & 0.69 & 0.19 & 0.21 & 0.48 & 0.62 & 0.24 & 0.15 & 0.59 & 0.77 \\
\textbf{T-locality }   & 0.14 & 0.33 & 0.71 & 0.03 & 0.27 & 0.48 & 0.77 & 0.09 & 0.45 & 0.57 & 0.55 & 0.07 & 0.57 & 0.67 & 0.63 & 0.11 \\
\hline
\textbf{M-generality}  & 0.04 & 0.06 & 0.37 & 0.55 & 0.17 & 0.07 & 0.53 & 0.58 & 0.08 & 0.12 & 0.16 & 0.39 & 0.14 & 0.13 & 0.29 & 0.46 \\
\textbf{M-locality}    & 0.61 & 0.71 & 0.70 & 0.54 & 0.59 & 0.88 & 0.67 & 0.57 & 0.77 & 0.87 & 0.79 & 0.49 & 0.87 & 0.88 & 0.77 & 0.57 \\
\hline
\textbf{C-generality} & 0.04 & 0.06 & 0.23 & 0.68 & 0.08 & 0.26 & 0.25 & 0.51 & 0.06 & 0.23 & 0.31 & 0.34 & 0.10 & 0.16 & 0.22 & 0.39 \\
\textbf{C-locality}    & 0.02 & 0.01 & 0.47 & 0.35 & 0.04 & 0.09 & 0.14 & 0.23 & 0.11 & 0.09 & 0.47 & 0.29 & 0.16 & 0.15 & 0.38 & 0.15 \\
\hline
\textbf{Temporality}   & 0.48 & 0.55 & 0.45 & 0.58 & 0.56 & 0.61 & 0.60 & 0.72 & 0.56 & 0.61 & 0.60 & 0.57 & 0.57 & 0.59 & 0.47 & 0.62 \\
\midrule
\textbf{Overall}       & 0.07 & 0.07 & \textbf{0.46} & \underline{0.21} & 0.14 & 0.23 & \textbf{0.39 }& \underline{0.36} & 0.16 & 0.25 & \textbf{0.39} & \underline{0.30} & 0.23 & 0.27 & \textbf{0.45} & \underline{0.35} \\
\bottomrule
\end{tabular}%
}
\end{table}

\begin{figure}[t]
    \centering
    \includegraphics[width=1\linewidth]{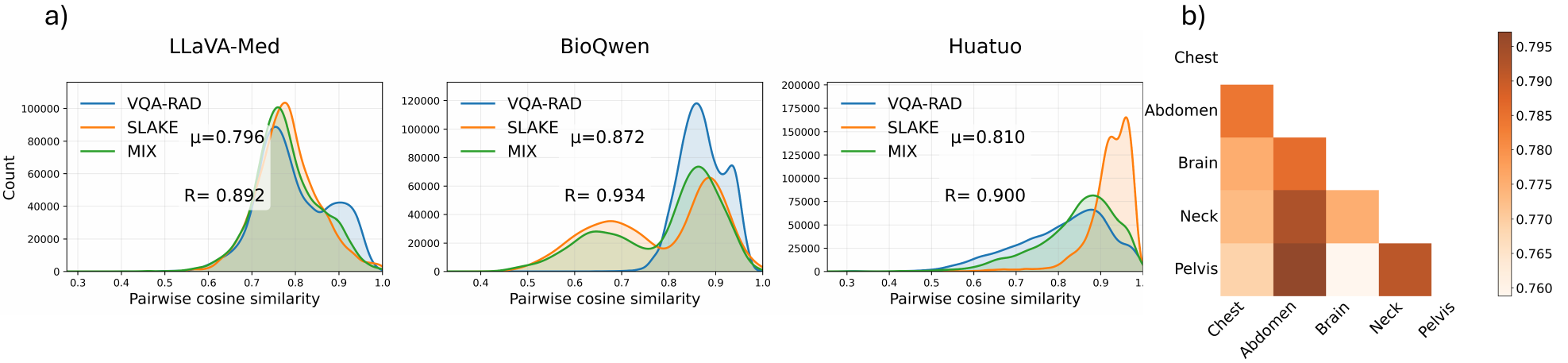}
    \vspace{-24pt}
    \caption{(\textbf{a}) \textbf{Histograms of the cosine similarity} between 1.5M pairs of embeddings across datasets. The average cosine similarity and mean resultant length are high and the minimum is above 0, indicating that the embedding space is a narrow cone. The same phenomenon exists for the pooled and the last hidden state. 
    (\textbf{b}) \textbf{Heatmap of cosine similarity} for 5 different anatomic sites (500 image-text pairs each) in LLaVA-Med displays high similarity. }
    \vspace{-18pt}
    \label{fig:cone}
\end{figure}

\vspace{-10pt}
\subsection{Cone Effect in Medical VLM Representations} 
\label{subsec:cone}
Before any editing, we first examine the baseline geometry of pre-trained medical VLM representations. To understand how distinct medical concepts are distributed within the latent space, we analyze the embeddings of 4,000 image-text pairs sampled from the SLAKE and VQA-RAD datasets using three different VLMs. Across all backbones, we observe that these representations are profoundly anisotropic, concentrating tightly around a narrow cone. Fig.~\ref{fig:cone}(a) visualizes this phenomenon by plotting the distribution of pairwise cosine similarities computed from 1.5 million randomly sampled embedding pairs. This "cone effect" is a known phenomenon where deep neural network representation clusters into a tightly constrained region rather than occupying the full available embedding space~\cite{liang2022mind}. To quantify this severity, we measure the mean resultant length, $R = \|\frac{1}{N}\sum_{i=1}^N x_i\|$, where $x_i$ are L2-normalized embeddings. In directional statistics, $R$ captures how strongly vectors pull toward a single mean direction: an $R$ score approaching 1.0 dictates that all embeddings point in nearly the exact same direction, whereas $R = 0$ indicates a perfectly uniform, isotropic distribution. A mean cosine similarity around 0.8 and $R$ around 0.9 restrict the effective representation space to a microscopic cone on the hypersphere. As shown in Fig.~\ref{fig:cone}b, this geometric concentration forces 5 disparate medical topics within LLaVA-Med to have astonishingly high similarity. This observation yields two critical implications for clinical VLM editing. First, because concepts are crowded into this narrow cone, localized edits are intrinsically difficult. A perturbation intended to correct one specific disease concept may easily intrude its geometric "neighbors". Second, cone tightness differs significantly across VLM backbones (Fig.~\ref{fig:cone}), suggesting that editing hyperparameters, especially those governing edit decision thresholds or key separation in memory-based editors, will not transfer reliably across models. We revisit this backbone dependency in Sec.~\ref{subsec:alpha}.

\subsection{Non-target Concept Drift Explains Locality Failures} We then connect representation geometry to the empirical locality gap between gradient-based and memory-based editors. Intuitively, if concept embeddings are tightly packed in representation space (Sec.~\ref{subsec:cone}), editing methods that induce broader changes around a target concept are more prone to break nearby, non-target concepts.

\begin{figure}[t]
    \centering

    \resizebox{\linewidth}{0.85\height}{\includegraphics[width=1\linewidth]{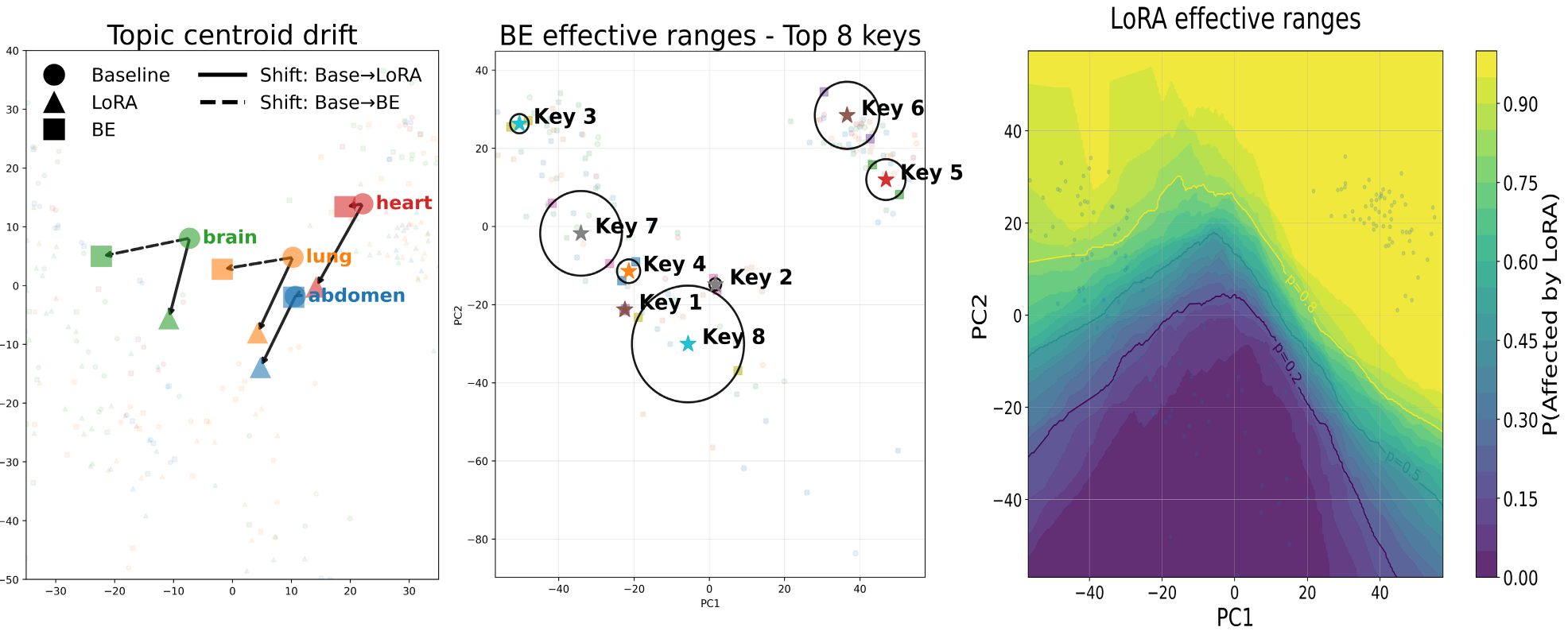}}
    \vspace{-18pt}
    \caption{\textbf{Left:} LoRA edits move all topic centroids of intended (Brain and Lung) and unintended targets (Abdomen and Heart), where BE restricts movement almost entirely to the edited topics. \textbf{Middle:} BE strictly restricts it edit activation regions by setting a binary decision sphere. 
    \textbf{Right:} LoRA fix targeted errors by broadly warping representations, and this global reshaping inevitably collides with neighboring clinical concepts.}
    \vspace{-18pt}
    \label{fig:drift_pca}
\end{figure}

To study this, we construct a simplified subset with 4 disease topics, including brain, lung, abdomen, and heart. We perform sequential edits only on brain and lung and evaluate on all 4 topics on Huatuo-7B. Under ideal locality, only brain and lung samples should change, while abdomen and heart remain unchanged. Fig.~\ref{fig:drift_pca} (left) plots these embeddings in a shared PCA basis. Centroid shift vectors reveal that LoRA moves the centroids of all four topics, while BalancEdit restricts movement almost entirely to the edited topics, directly reflecting the observed locality gap. We simulate and visualize the effective ranges of both editors in the PCA space to explain these drift patterns. Fig.~\ref{fig:drift_pca} (middle) illustrates BalancEdit’s key-query matching, where localized spherical boundaries around keys formed during training determine if a test sample triggers an edit. This induces binary, spatially restricted behavior: samples outside these boundaries do not activate the edit, protecting non-target regions. Conversely, Fig.~\ref{fig:drift_pca} (right) maps LoRA’s effective range to a spatial contour by tracking how often LoRA changes the model’s output over 100 independent LoRA runs. The resulting broad contours demonstrate that LoRA's distributed weight updates continuously alter the embedding manifold, leading to widespread non-target concept drift.

\subsection{Cases Where Memory-based Editors Fail} 
\begin{figure}[t]
    \centering
    \includegraphics[width=1\linewidth]{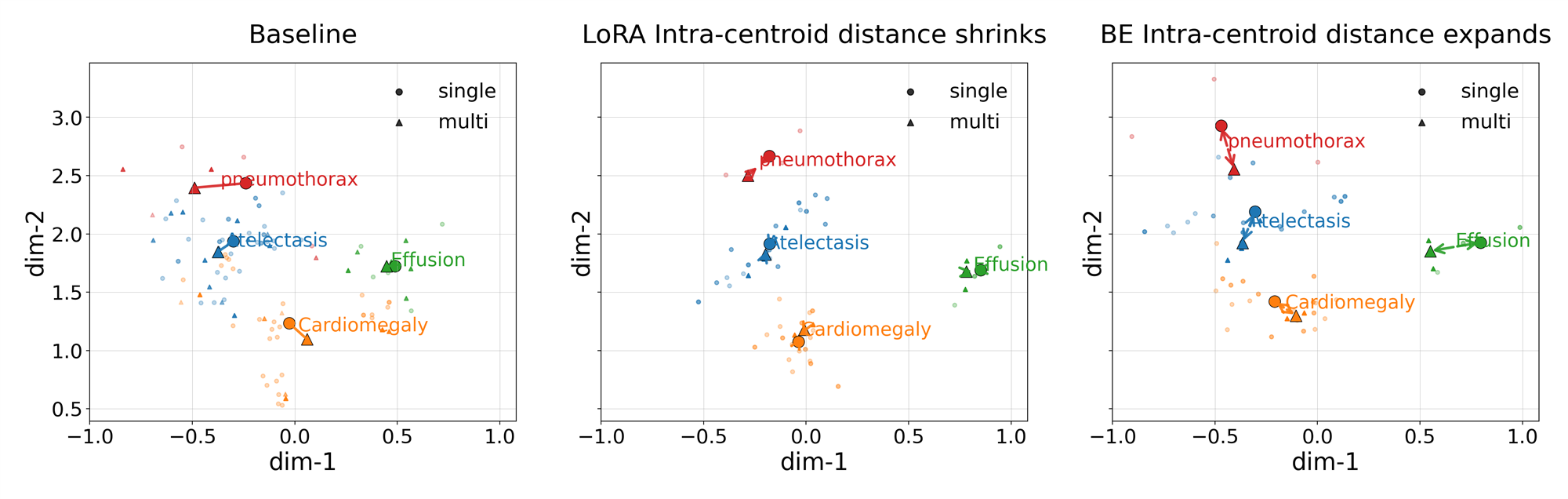}
    \vspace{-18pt}
    \caption{\textbf{Composition probe: single- vs.\ multi-finding geometry under editing.} PCA visualization of embeddings for four diseases, where circles denote single-finding cases and triangles denote multi-finding cases. Single and multi cases heavily overlap pre-edit, indicating concept entanglement. LoRA tends to reduce the single--multi centroid gap, whereas BalancEdit often increases the gap.}
    \label{fig:comp_pca}
    \vspace{-18pt}
\end{figure}

Next, we investigate why BE fails in clinical composition tasks and shows relatively lower generality. We construct a simplified task that takes 4 diseases: pneumothorax, effusion, cardiomegaly, and atelectasis. For each disease, we split samples into single-label finding samples (where the patient has this disease only) and multi-label finding samples (where the patient has other co-occurring diseases). We edit using only the single-finding subset and evaluate their transfer to multi-label finding samples.
Fig.~\ref{fig:comp_pca} visualizes embeddings for single (circles) versus multi (triangles) contexts (left: base, middle: LoRA, right: BalancEdit). Notably, single and multi points overlap heavily in the baseline space, suggesting that composition tasks often correspond to subtle, entangled latent space clusters that can not be isolated by a simple binary spherical boundary. For each disease, LoRA reduces the distance between the centroids of single-finding samples and those of multi-finding centroids. On the other hand, BalancEdit pushes them farther apart. This aligns with Tab.\ref{tab:seqresults}: BE tends to under-transfer from single-finding edits to clinically equivalent multi-finding contexts, yielding lower generality.
This study indicates one drawback of memory editors’ reliance on a binary activation radius: the edit is applied only when a test sample falls inside a learned activation region, making it difficult to compose partially overlapping evidence from multiple findings. In a space where different concepts, such as single- and multi-finding samples, already interleave, such hard gating can either miss clinically equivalent multi-finding samples or require expanding the region. 

\subsection{Where and What to Edit: Sensitivity to Components and Layers}
 A practical question in multimodal editing is \emph{where} to intervene inside a VLM under strict efficiency constraints. Although different editing methods can have comparable parameter footprints, they often operate at very different architectural depths and exhibit markedly different behaviors. For example, in BioQwen, applying LoRA across all 27 language MLP layers (with rank $r{=}16$) introduces roughly 18M trainable parameters, whereas BalancEdit modifies the final MLP layer’s up\_projection with 13.8M trainable parameters. This raises two natural questions: i) given a similar parameter budget, does LoRA’s stronger generality come from distributing small updates across many layers, and ii) can we combine LoRA and BalancEdit to capture complementary benefits?
 
In this study, we use one hybrid design, which we refer to as \emph{BELoRA}. Concretely, we keep BalancEdit’s key-based activation mechanism, but parameterize each editable module by a LoRA adapter rather than a full weight rewrite. When a test sample is deemed eligible for editing, the corresponding LoRA adapter is activated; otherwise, the model falls back to its original computation. This design keeps edits parameter-efficient while allowing us to control \emph{where} and \emph{how deep} the intervention is applied.

Beyond depth, VLMs expose multiple editable components, including the language model (LM), the vision encoder, and the vision--language projector. However, most existing editing benchmarks default to intervening only in the final LM layer, leaving the third question iii) unanswered: what is the impact of editing different multimodal components?

\begin{figure}[t]
 \centering
 \includegraphics[width=1\linewidth]{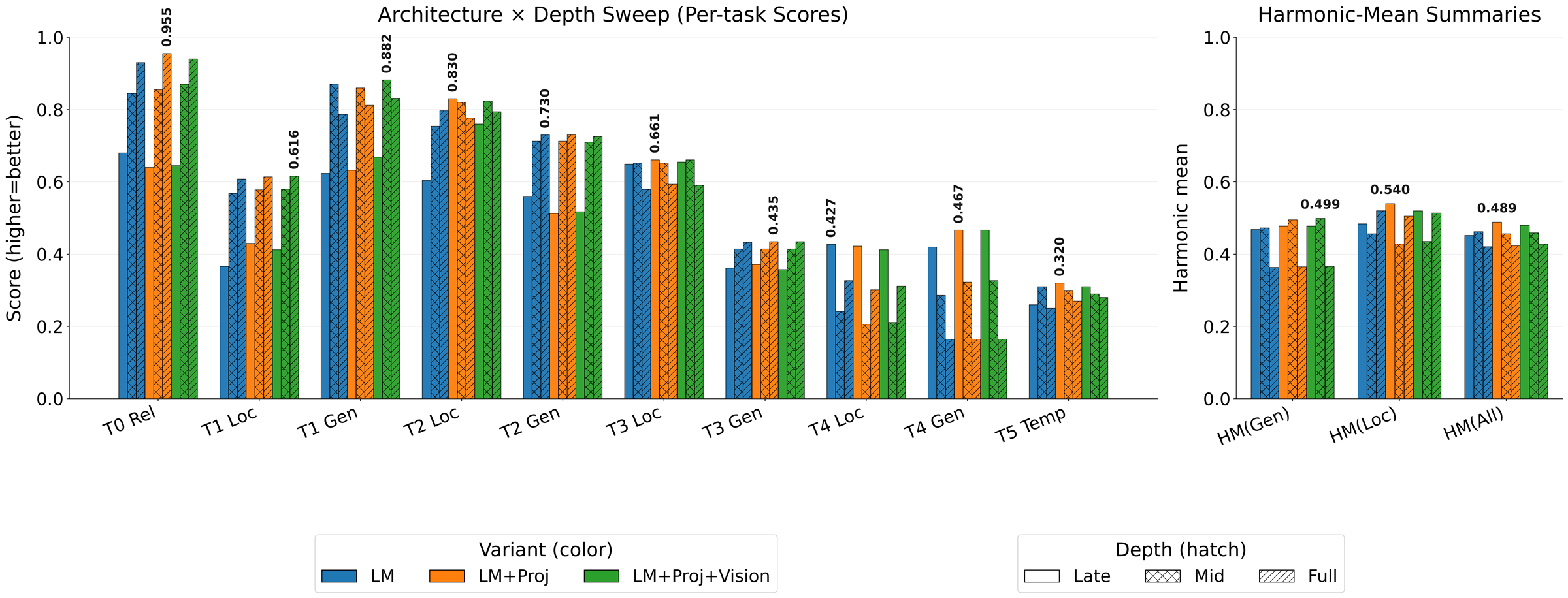}
 \vspace{-16pt}
 \caption{\textbf{Architecture sweep using \emph{BELoRA} over varying model components and depths.} 
 \textbf{Left:} Task-wise results. 
 \textbf{Right:} Harmonic mean of generality tasks, locality tasks, and all tasks.}
 \vspace{-15pt}
 \label{fig:archi_sweep}
 \end{figure}

To answer these three questions, we conduct a controlled ablation study on Huatuo-7B using \emph{BELoRA} and vary the LM depth of intervention (Late, Late+Mid, Full) and the editable scope (LM; LM+projector; LM+projector+vision encoder). As shown in Fig.~\ref{fig:archi_sweep}, we summarize performance using harmonic means over locality tasks, generality tasks, and all tasks.
Overall, we find that while a hybrid \emph{BELoRA} outperforms both BE and LoRA, it still struggles on compositionality and temporality tasks. Counterintuitively, expanding the hybrid across more layers does not linearly improve generality; instead, late+mid-layer interventions tend to favor generality, while late-layer interventions favor locality and yield the best overall balance. Finally, regarding component choice beyond the LM, editing the vision--language projector and late LM layers provides benefits with little extra computation; in contrast, expanding the edit scope further to include the vision encoder provides no consistent gains and reduces stability.

\subsection{Backbone-Dependent Hyperparameter Sensitivity in Memory-Based Editors}

\label{subsec:alpha}

\begin{figure}[t]
    \centering
    \resizebox{\linewidth}{5.5cm}{\includegraphics{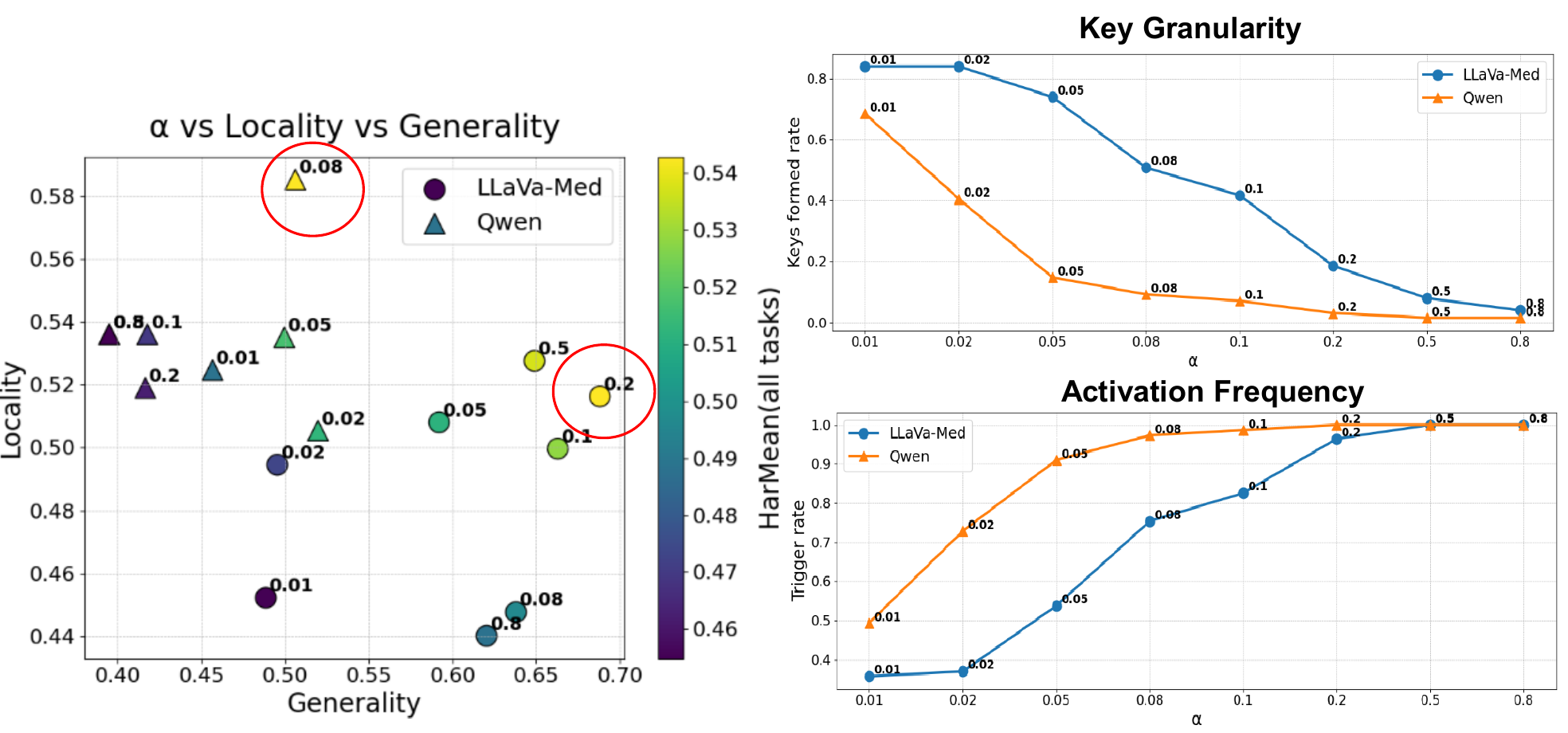}}
    \vspace{-16pt}
    \caption{\textbf{Left:} Different backbones have different optimal points (highest overall score marked by red circle), and the same hyperparameter may have opposite effects in different backbones. 
    \textbf{Right:} Extremely low or high radius result in degraded performance.}
     \vspace{-18pt}
    \label{fig:alpha_sweep}
\end{figure}

Memory-based editors such as BalancEdit store edit inputs and their corresponding edited parameters as key--value pairs during training and reuse them at inference time by matching test samples to stored keys. A central hyperparameter is the \textbf{radius} $\alpha$, which defines an acceptance region around each key: a test sample triggers an edit if its embedding falls within $\alpha$ of at least one key. Crucially, $\alpha$ controls two coupled behaviors: (i) \textbf{key granularity} during construction (whether training edits form many fine keys or merge into fewer coarse keys), and (ii) \textbf{activation frequency} at test time (how often edits are applied). Sweeping $\alpha$ on BioQwen and LLaVA-Med reveals two consistent failure patterns (Fig.~\ref{fig:alpha_sweep}, right). When $\alpha$ is too small, the editor forms many keys but activates them rarely, leading to under-correction. When $\alpha$ is too large, keys merge aggressively and activation becomes frequent, but merged keys mix disparate concepts and introduce interference, degrading locality (and often generality). Thus, $\alpha$ is highly critical to editing success.

However, we observe the optimal operating range of $\alpha$ is highly backbone-dependent. Fig.~\ref{fig:alpha_sweep} (left) shows that the best locality--generality trade-off occurs vastly different for BioQwen than LLaVA-Med; a radius that is well-calibrated for one VLM can therefore place another in either the under-activation failure mode or the over-mixing failure mode, yielding avoidable performance loss.

We trace this sensitivity back to differences in representation geometry. As indicated by the cone effect (Fig.~\ref{fig:cone}), backbones differ in how angularly concentrated their embeddings are. In more concentrated spaces like BioQwen’s, small changes in $\alpha$ can disproportionately change which samples fall inside key neighborhoods, simultaneously altering key merging during construction and edit activation at inference. Consequently, $\alpha$ (and activation radius, more broadly) is not easily transferable across backbones, and robust use of memory-based editors requires careful calibration.

\subsection{Case Study}
\begin{figure}
    \centering
    \includegraphics[width=1\linewidth]{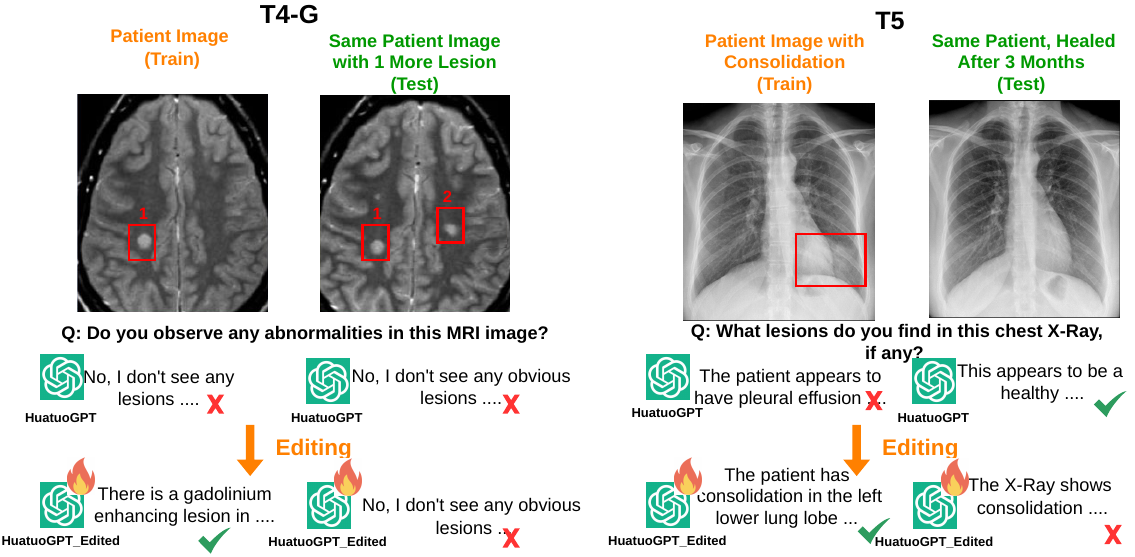}
    \caption{BE Failed Cases for T4-G and T5 on Huatuo34B.}
    
    \label{fig:cases}
\end{figure}

Fig~\ref{fig:cases} shows two failure cases of BalancEdit. T4-G (Left): an edit successfully teaches the model to identify a single brain lesion but fails to transfer to the same patient's image with one additional lesion present. We attribute this to the binary-gating limitation discussed above. T5 (Right): an edit correcting a baseline consolidation diagnosis propagates incorrectly to a healthy three-month follow-up, where the edited model asserts consolidation on a healthy study overriding a previously correct longitudinal judgment. These cases make concrete why compositional and temporal axes remain unsolved by current editors.
\section{Conclusion}
\label{sec:concl}
We introduced \ours~to evaluate medical VLM editing under realistic clinical challenges. Our findings reveal key trade-offs: gradient-based methods generalize well but degrade locality, while memory-based editors preserve locality but struggle with compositional transfer and hyperparameter sensitivity. Our geometric analysis provides explanations of these discrepancies and highlights potential drawbacks of existing methods, motivating future methodological improvements.
\section{Acknowledgements}
We acknowledge funding support by National Science Foundation (NSF) via grant IIS-2435746, Defense Advanced Research Projects Agency (DARPA) under contract No. HR00112520042, as well as the University of Michigan MIDAS PODS Grant Award.
\bibliographystyle{splncs04}
\bibliography{main}

\appendix

\section{Benchmark Dataset Details}

\subsection{Data sources.}
The benchmark integrates samples from public medical datasets including \textbf{VQA-RAD~\cite{lau2018dataset}}, \textbf{PMC-VQA~\cite{zhang2023pmc}}, \textbf{PadChest-GR~\cite{de2025padchest}}, and \textbf{SLAKE~\cite{liu2021slake}}. Per-task source proportions differ depending on whether a task requires patient-level linkage for temporal evaluation, cross-modality alignment for shift evaluation, or semantic matching for generality evaluation. We report the exact composition of each task in the corresponding task subsections.

\subsection{Data Construction}
\label{app:distillation}
Our construction pipeline follows a two-stage distillation process that transforms heterogeneous supervision into controlled evaluation sets.

\paragraph{Stage 1: Clinical Attribute Distillation (QA $\rightarrow$ Fact).}
For each (image, question, answer) instance, we use a large language model (LLM) to extract a concise standardized medical fact statement from the QA pair. The LLM is prompted with the image context (when available), the question, and the ground-truth answer, and outputs a short fact sentence together with a structured attribute list. The attribute list includes: (i) condition or finding, (ii) anatomical site, (iii) imaging modality, and when available, acquisition/view/protocol cues, (iv) question type, and (v) progression-related attributes for longitudinal settings. This step reduces annotation noise from phrasing and formatting differences and enables more consistent alignment across data sources.

\paragraph{Stage 2: Image-level Profiling and Evaluation Set Assembly.}
Because many datasets contain multiple questions per image, we aggregate all distilled facts belonging to the same image into an \emph{image profile}. An image profile is a compact set of structured attributes and concepts associated with that image, obtained by merging and summarizing per-question outputs to remove redundancy and resolve minor inconsistencies. These profiles enable programmatic, controlled sampling for benchmark construction: we can isolate a specific clinical variable while holding others fixed, and generate paired or multi-image evaluation sets that stress different desiderata of model editing. For example, we can construct locality-oriented sets where surface overlap exists but edits should not propagate (e.g., same template but different diseases), and generality-oriented sets where clinically similar cases differ along one axis (e.g., same finding across different images, or protocol/view shifts). The resulting evaluation sets are serialized into JSONL/CSV formats with explicit grouping metadata for downstream editing and evaluation.
\subsection{Task Construction Protocol}
\label{app:task-construct}
Using the image profiles, we construct benchmark tasks by instantiating controlled \textbf{editing instances}. Each task instance consists of (i) an \textbf{edit request}, namely a target question with a corrected answer/fact to be edited into the model, and (ii) a set of \textbf{probe questions} used to evaluate post-edit behavior such as reliability, locality, and generality. Thus, the benchmark is organized around edit instances rather than standalone images or raw QA pairs. For example, locality probes may keep the question template fixed while swapping in clinically unrelated images, whereas generality probes may use clinically matched images, paraphrased questions, modality-shifted counterparts, or temporal follow-up pairs. Each task instance is stored with annotations of (i) identifiers for the edit target and probes, (ii) the structured attributes used for matching, and (iii) the desired behavior under successful editing (efficacy on the edit target, minimal side effects on locality probes, and appropriate transfer on generality probes).

\subsection{Data Distribution of the Benchmark}
\label{app:data_distribution}

\paragraph{Overall benchmark composition.}
Our benchmark is built on a unified pool of \textbf{1398 unique images} paired with \textbf{16,276 questions (edit request questions + probe questions)}. 

\begin{figure}[t]
    \centering
    \includegraphics[width=0.95\linewidth]{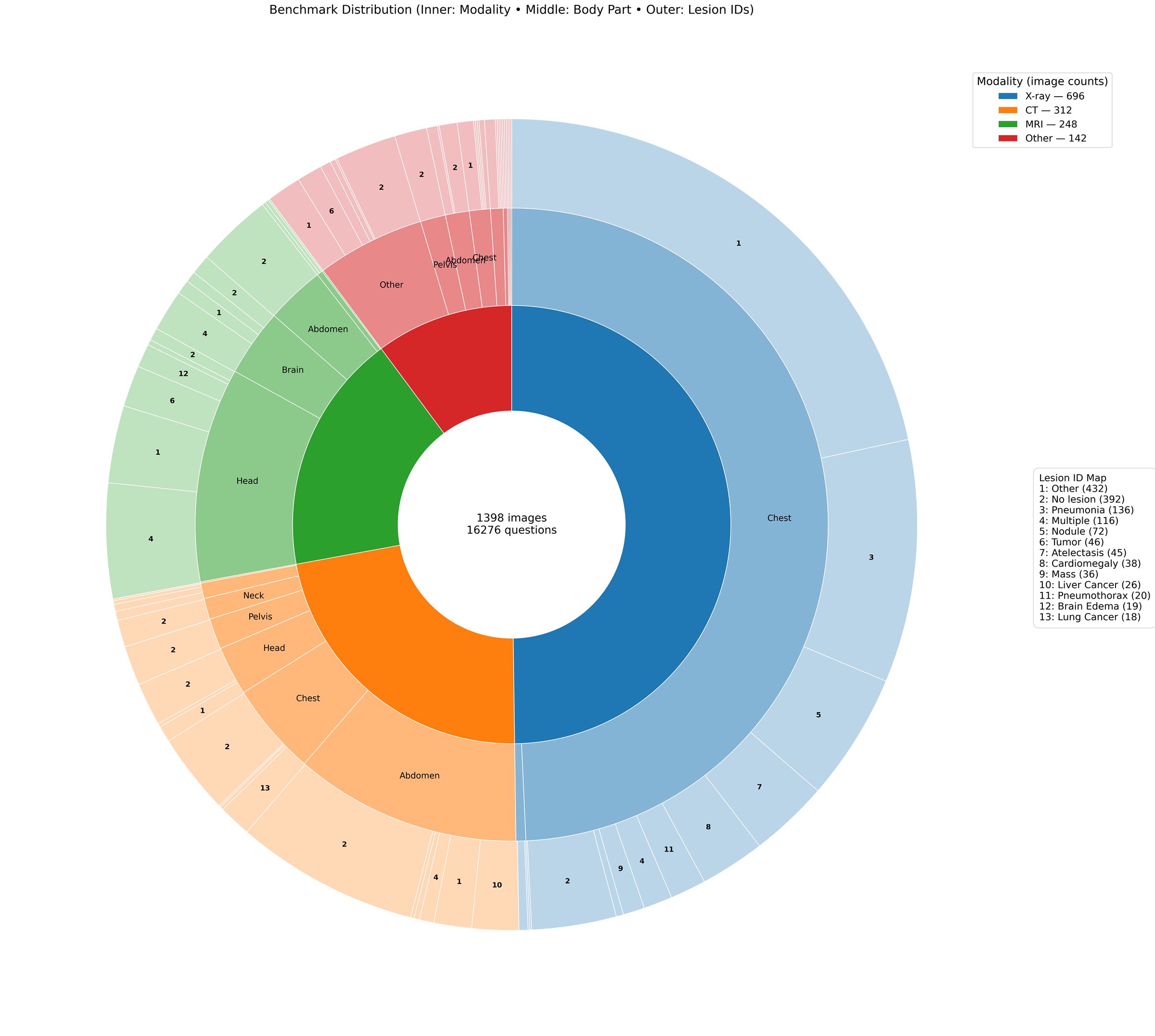}
    \caption{\textbf{Benchmark distribution} (inner ring: modality; middle ring: body part; outer ring: lesion IDs). Center reports global counts (\#images, \#questions).}
    \label{fig:dataset_sunburst}
\end{figure}

Figure~\ref{fig:dataset_sunburst} summarizes the global distribution using a concentric sunburst diagram (inner: modality, middle: anatomic site, outer: lesion ID).

\paragraph{Modalities and topic coverage.}

Across the full dataset, modality counts (by images) are: CT 312, X-ray 696, MRI 248, and Other 142.
To quantify clinical topic coverage, we map each image-question pair to a lesion-ID label set. The global lesion ID map shows a large fraction of samples are labeled as No lesion (392) or Other (432), followed by multi-label cases Multiple (116). Among specific lesion categories, the most frequent include Tumor (46), Mass (36), Cardiomegaly (38), Pneumonia (136), Liver Cancer (26), Atelectasis (25), Pneumothorax (20), Brain Edema (19), Lung Cancer (18), Nodule (72), and Pulmonary Mass (10) (see the lesion ID map in Figure~\ref{fig:dataset_sunburst}).

\paragraph{Task-wise distribution.}
Each evaluation task probes a specific dimension of multimodal model editing (reliability, locality, generality, and robustness under shifts). When tasks are formed by sampling from the base pool or by applying controlled shifts to base examples, they largely preserve the base dataset’s clinical distribution, while introducing task-specific structures such as prompt-sets, paired shifts, or compositional counterparts.

\medskip

\noindent\textbf{T0: Reliability on Edited Errors.}
T0 is formed by sampling originally incorrect cases from the base dataset as edit instances. Each edit instance consists of a single edit request without additional probes, and tests whether the editor successfully corrects the target error. In the main experiments, we use 200 edit instances, which provides a consistent evaluation budget across methods and backbones.

\noindent\textbf{T1L: Same Question, Different Image (Locality).}
T1L tests whether an edit breaks unrelated images under the same prompt. The T1L evaluation set contains 968 edit instances spanning 2 sources (SLAKE 832, VQA-RAD 136). Each edit instance consists of one edit request, defined by an image-question pair, together with a set of locality probes formed by pairing the same question with multiple different images. The number of images per question is mean 14.38, median 5, min 2, and max 351. Answer diversity (unique answers per question) is mean 2.32 and median 2, and 16 question groups contain duplicated image IDs as sanity checks. Aggregating answers across all T1L probe images, the most frequent labels include No (1285) and Yes (1247), along with common clinical tokens such as CT, MRI, X-Ray, Lung, and Chest. The largest prompt groups (by number of images) include ``Which part of the body does this image belong to?'' (351) and ``What modality is used to take this image?'' (321).

\medskip
\noindent\textbf{T1G: Same Question, Semantically Matched Image (Generality).}
T1G is constructed via an image-level shift applied to error-derived anchors while keeping the question template fixed. Each edit instance consists of one edit request together with probe questions formed by pairing the same question with clinically matched images that share the target semantics. These matched images preserve the target finding while varying the specific patient image, allowing us to test whether the edit transfers beyond the original instance.

\medskip
\noindent\textbf{T2L: Same Image Context, Different Question (Locality).}
We instantiate T2L using a cross-image disease-consistency set of 685 edit instances. Each edit instance consists of one edit request together with 2--3 probe image-question-answer tuples that share the same disease label as the edit request but use different questions. This task evaluates whether an edit spills over to related disease contexts under different questions.

\medskip
\noindent\textbf{T2G: Same Image, Semantically Equivalent Text (Generality).}
T2G is constructed via a text-level shift applied to error-derived anchors while keeping the image fixed. Each edit instance consists of one edit request together with a set of probe questions that are semantic-preserving rewrites of the original question, including paraphrases, synonym substitutions, or alternative clinical phrasing. This task evaluates whether the edit generalizes across equivalent textual formulations of the same clinical query~\cite{cai2025does,koopman2023dr}.

\medskip
\noindent\textbf{T3L/T3G: Modality Shift (A$\rightarrow$B).}
We instantiate the modality-shift tasks (T3L/T3G) using a cross-modality same-disease set containing 113 edit instances with modality distribution X-ray 46, MRI 42, and CT 25. Each edit instance consists of one edit request together with a set of probe image-question-answer tuples drawn from matched images in other modalities but sharing the same disease context. Each anchor is associated with a small disease set (mean 1.080, median 1) and multiple matched images in other modalities (mean 8.850, median 8). This task evaluates whether the edit remains appropriately specific under acquisition and modality shifts: T3G measures transfer to pre-wrong matched cases, while T3L measures preservation on pre-correct matched cases.

\medskip
\noindent\textbf{T4L/T4G: Compositional Consistency.}
We instantiate the compositional tasks using two subsets. For T4L, the compositional-locality subset contains 257 edit instances across 116 unique images. Each edit instance consists of one edit request on a target finding together with a probe question about a different co-occurring finding in the same image, testing whether the edit disrupts other correct concepts within a compositional case. For T4G, the compositional-generality subset contains 1000 edit instances spanning 24 unique single-lesion labels and 36 unique multi-lesion combinations. Each edit instance consists of one edit request defined on a single-finding case together with probe questions on multi-finding cases that contain the edited concept. Brain Edema is the most frequent single-lesion label (660), followed by tumor (63) and mass (61). Together, these tasks test whether edits transfer to compositional clinical settings while preserving unrelated co-occurring findings.

\medskip
\noindent\textbf{T5: Temporal Consistency (Same Patient, Different Timepoints).}
Clinical deployment frequently involves longitudinal reasoning: a patient returns for a follow-up visit, and clinicians compare the current study with a prior study to determine whether a finding is persistent, new, or resolved. These follow-up images are often very similar, with only subtle changes that reflect disease progression, recovery, or stabilization. A successful edit should therefore avoid confusing one visit for the same patient with another. We instantiate T5 with 324 edit instances derived from 440 curated images forming prior--current visit pairs, spanning 43 diseases from PadChest-GR. We encode progression into a binary presence code, where stable/improving/worsening map to “11” (present in both visits), new maps to “01” (absent→present), and resolved maps to “10” (present→absent). For each pair, we edit on the prior visit image with a templated presence question and evaluate transfer on the corresponding current image using an analogous question.

\subsection{Task Score Calculation}
Each edit request is $e=(I^\star,q^\star,y^\star)$ where the pre-edit model answers incorrectly, $f_\theta(I^\star,q^\star)\neq y^\star$, and $y^\star$ is the clinician-desired answer. Let the edited model be $f_{\theta'}=\mathcal{A}(f_\theta;e)$. For any example $(I,q,y)$, define pre- and post-edit correctness
$c_{\mathrm{pre}}(I,q,y)=\mathbf{1}[f_{\theta}(I,q)=y]$ and
$c_{\mathrm{post}}(I,q,y)=\mathbf{1}[f_{\theta'}(I,q)=y]$.
We measure \textbf{locality} by the flip rate (breakage among pre-correct examples),
\[
\mathrm{Flip}(\mathcal{D};e)=
\frac{\sum_{(I,q,y)\in\mathcal{D}} c_{\mathrm{pre}}(I,q,y)\,(1-c_{\mathrm{post}}(I,q,y))}
{\sum_{(I,q,y)\in\mathcal{D}} c_{\mathrm{pre}}(I,q,y)+\epsilon},
\]
and \textbf{generality} by the fix rate (repair among pre-wrong examples),
\[
\mathrm{Fix}(\mathcal{D};e)=
\frac{\sum_{(I,q,y)\in\mathcal{D}} (1-c_{\mathrm{pre}}(I,q,y))\,c_{\mathrm{post}}(I,q,y)}
{\sum_{(I,q,y)\in\mathcal{D}} (1-c_{\mathrm{pre}}(I,q,y))+\epsilon}.
\]
We report locality as $1-\mathrm{Flip}$ so that higher is better. Metrics are macro-averaged over edit requests $\mathcal{E}$:
$\mathrm{Metric}_t=\frac{1}{|\mathcal{E}|}\sum_{e\in\mathcal{E}}\mathrm{Metric}_t(e)$.

\paragraph{T0: Reliability (target correction).}
An edit is only meaningful if it corrects the observed failure on the target input. We set $\mathcal{D}_0(e)=\{(I^\star,q^\star,y^\star)\}$ and report post-edit correctness $c_{\mathrm{post}}(I^\star,q^\star,y^\star)$.

\paragraph{T1L: Image Locality (same question, different image).}
Clinicians reuse templated questions across patients; a failure mode is memorizing the prompt while ignoring visual evidence. We keep $q^\star$ fixed and pair it with unrelated images $\tilde I$ (from different condition/site groups), forming
$\mathcal{D}_{1\mathrm{L}}(e)=\{(\tilde I,q^\star,\tilde y)\}$ where $\tilde y$ is the ground-truth answer for $(\tilde I,q^\star)$. We report $1-\mathrm{Flip}(\mathcal{D}_{1\mathrm{L}}(e);e)$.

\paragraph{T1G: Image Generality (same question, same-semantic image).}
The same finding recurs across patients; an ideal edit should transfer beyond $I^\star$. We keep $q^\star$ fixed and replace $I^\star$ with the same-semantic images $I^{\mathrm{sem}}$ (matched on condition/site and relevant metadata when available), forming
$\mathcal{D}_{1\mathrm{G}}(e)=\{(I^{\mathrm{sem}},q^\star,y^{\mathrm{sem}})\}$.
We report $\mathrm{Fix}(\mathcal{D}_{1\mathrm{G}}(e);e)$.

\paragraph{T2L: Text Locality (same image context, different question).}
Edits must preserve other correct behaviors on the same study (e.g., other findings in the same scan). We keep $I^\star$ fixed and query a different clinically relevant question $q^{\mathrm{alt}}$ with pre-edit correctness $c_{\mathrm{pre}}=1$, forming
$\mathcal{D}_{2\mathrm{L}}(e)=\{(I^\star,q^{\mathrm{alt}},y^{\mathrm{alt}})\}$,
and report $1-\mathrm{Flip}(\mathcal{D}_{2\mathrm{L}}(e);e)$.

\paragraph{T2G: Text Generality (semantic-preserving rewrites).}
Medical language varies widely (abbreviations, shorthand, formatting, negation, reasoning styles). We keep $I^\star$ fixed and evaluate four semantic-preserving rewrites $\{q^{(k)}\}_{k=1}^4$ of $q^\star$ (paraphrase; synonym/abbreviation; question-form change; telegraphic clinical note style), forming
$\mathcal{D}_{2\mathrm{G}}(e)=\{(I^\star,q^{(k)},y^\star)\}_{k=1}^4$,
and report $\mathrm{Fix}(\mathcal{D}_{2\mathrm{G}}(e);e)$.

\paragraph{T3L/T3G: Acquisition \& Modality Shift ($A\rightarrow B$).}
Findings can appear differently across scanners, protocols, views, or modalities. We edit on acquisition setting $A$ for concept $x$, then evaluate concept-matched cases under $B$, split by pre-edit correctness:
\[
\begin{aligned}
\mathcal{D}_{3\mathrm{G}}(e) &= \{(I^{B},q^{B},y^{B}) : x,\ c_{\mathrm{pre}}(I^{B},q^{B},y^{B})=0\},\\
\mathcal{D}_{3\mathrm{L}}(e) &= \{(I^{B},q^{B},y^{B}) : x,\ c_{\mathrm{pre}}(I^{B},q^{B},y^{B})=1\}.
\end{aligned}
\]
We report $\mathrm{Fix}(\mathcal{D}_{3\mathrm{G}}(e);e)$ and $1-\mathrm{Flip}(\mathcal{D}_{3\mathrm{L}}(e);e)$.

\paragraph{T4L/T4G: Clinical Composition (co-occurring findings).}
Real cases are compositional, and clinicians ask multiple questions per image. We probe (i) \emph{Compositional generality:} edit on a single-finding instance and evaluate on multi-finding cases that contain the edited finding, forming $\mathcal{D}_{4\mathrm{G}}(e)$; and (ii) \emph{Compositional locality:} on the same multi-finding image $I^\star$, the target question is pre-wrong while an auxiliary question about another finding $q^{\mathrm{cons}}$ is pre-correct, forming $\mathcal{D}_{4\mathrm{L}}(e)=\{(I^\star,q^{\mathrm{cons}},y^{\mathrm{cons}})\}$.
We report $\mathrm{Fix}(\mathcal{D}_{4\mathrm{G}}(e);e)$ and $1-\mathrm{Flip}(\mathcal{D}_{4\mathrm{L}}(e);e)$.

\paragraph{T5: Temporal Consistency (prior $\rightarrow$ follow-up).}
Edits should not introduce contradictions across longitudinal follow-ups where progression/regression matters. We edit on the earlier timepoint $(I_{t_0},q^\star,y^\star)$ and evaluate the paired follow-up at $t_1$, forming $\mathcal{D}_5(e)=\{(I_{t_1},q_{t_1},y_{t_1})\}$. We report 1 - percentage an edit at earlier time introduces factual errors in later follow-up studies.

\section{Additional Results}
\subsection{Full Experiments Table}
In the main text, we report the numerical table for sequential editing and show single-edit results as radar plots due to space constraints.

Here we include the full main results in Tabs.\ref{tab:llavamed7b_full_results_singlecol}, \ref{tab:bioqwen2b_full_results_singlecol}, \ref{tab:huatuo7b_full_results_singlecol}, \ref{tab:huatuo34b_full_results_singlecol}:

\begin{table}[t]
\centering
\scriptsize
\setlength{\tabcolsep}{3.5pt}
\resizebox{\columnwidth}{!}{%
\begin{tabular}{lcccccccc}
\toprule
\textbf{LLaVA-Med 7B} & \textbf{MEND} & \textbf{MEND-S} & \textbf{GRACE} & \textbf{GRACE-S} & \textbf{BE} & \textbf{BE-S} & \textbf{LoRA} & \textbf{LoRA-S} \\
\midrule
Reliability   & 0.26 & 0.11 & 0.57 & 0.33 & 1.00 & 0.60 & 1.00 & 0.96 \\
I-locality    & 0.39 & 0.18 & 0.47 & 0.28 & 0.59 & 0.42 & 0.40 & 0.46 \\
I-generality  & 0.29 & 0.12 & 0.68 & 0.43 & 0.82 & 0.63 & 0.98 & 0.95 \\
T-generality  & 0.33 & 0.27 & 0.47 & 0.39 & 0.71 & 0.59 & 0.73 & 0.71 \\
T-locality    & 0.52 & 0.14 & 0.47 & 0.33 & 0.88 & 0.71 & 0.65 & 0.03 \\
M-generality  & 0.05 & 0.04 & 0.01 & 0.06 & 0.05 & 0.37 & 0.22 & 0.55 \\
M-locality    & 0.69 & 0.61 & 0.93 & 0.71 & 0.98 & 0.70 & 0.95 & 0.54 \\
C-generality  & 0.17 & 0.04 & 0.09 & 0.06 & 0.31 & 0.23 & 0.63 & 0.68 \\
C-locality    & 0.07 & 0.02 & 0.04 & 0.01 & 0.29 & 0.47 & 0.29 & 0.35 \\
Temporality   & 0.53 & 0.48 & 0.66 & 0.55 & 0.55 & 0.45 & 0.42 & 0.58 \\
Overall       & 0.17 & 0.07 & 0.07 & 0.07 & 0.28 & 0.46 & 0.49 & 0.21 \\
\bottomrule
\end{tabular}%
}
\caption{Full experiment results on LLaVA-Med 7B. S denotes sequential editing.}
\label{tab:llavamed7b_full_results_singlecol}
\end{table}

\begin{table}[t]
\centering
\scriptsize
\setlength{\tabcolsep}{3.5pt}
\resizebox{\columnwidth}{!}{%
\begin{tabular}{lcccccccc}
\toprule
\textbf{BioMed-Qwen2-VL 2B} & \textbf{MEND} & \textbf{MEND-S} & \textbf{GRACE} & \textbf{GRACE-S} & \textbf{BE} & \textbf{BE-S} & \textbf{LoRA} & \textbf{LoRA-S} \\
\midrule
Reliability   & 0.24 & 0.09 & 0.66 & 0.51 & 1.00 & 0.72 & 1.00 & 0.98 \\
I-locality    & 0.50 & 0.63 & 0.75 & 0.71 & 0.61 & 0.40 & 0.63 & 0.63 \\
I-generality  & 0.31 & 0.37 & 0.73 & 0.61 & 0.77 & 0.42 & 1.00 & 0.95 \\
T-generality  & 0.43 & 0.27 & 0.17 & 0.22 & 0.79 & 0.53 & 0.79 & 0.69 \\
T-locality    & 0.69 & 0.27 & 0.70 & 0.48 & 0.95 & 0.77 & 0.90 & 0.09 \\
M-generality  & 0.09 & 0.17 & 0.05 & 0.07 & 0.22 & 0.53 & 0.41 & 0.58 \\
M-locality    & 0.79 & 0.59 & 0.96 & 0.88 & 0.72 & 0.67 & 0.94 & 0.57 \\
C-generality  & 0.13 & 0.08 & 0.21 & 0.26 & 0.23 & 0.25 & 0.33 & 0.51 \\
C-locality    & 0.04 & 0.04 & 0.07 & 0.09 & 0.18 & 0.14 & 0.18 & 0.23 \\
Temporality   & 0.58 & 0.56 & 0.63 & 0.61 & 0.48 & 0.60 & 0.48 & 0.72 \\
Overall       & 0.17 & 0.14 & 0.19 & 0.23 & 0.41 & 0.39 & 0.50 & 0.36 \\
\bottomrule
\end{tabular}%
}
\caption{Full experiment results on BioMed-Qwen2-VL 2B. S denotes sequential editing.}
\label{tab:bioqwen2b_full_results_singlecol}
\end{table}

\begin{table}[t]
\centering
\scriptsize
\setlength{\tabcolsep}{3.5pt}
\resizebox{\columnwidth}{!}{%
\begin{tabular}{lcccccccc}
\toprule
\textbf{Huatuo 7B} & \textbf{MEND} & \textbf{MEND-S} & \textbf{GRACE} & \textbf{GRACE-S} & \textbf{BE} & \textbf{BE-S} & \textbf{LoRA} & \textbf{LoRA-S} \\
\midrule
Reliability   & 0.18 & 0.13 & 0.73 & 0.54 & 1.00 & 0.67 & 1.00 & 0.58 \\
I-locality    & 0.45 & 0.38 & 0.56 & 0.31 & 0.64 & 0.34 & 0.62 & 0.59 \\
I-generality  & 0.46 & 0.47 & 0.68 & 0.52 & 0.73 & 0.39 & 0.95 & 0.69 \\
T-generality  & 0.38 & 0.19 & 0.24 & 0.21 & 0.52 & 0.48 & 0.73 & 0.62 \\
T-locality    & 0.68 & 0.45 & 0.77 & 0.57 & 0.76 & 0.55 & 0.81 & 0.07 \\
M-generality  & 0.11 & 0.08 & 0.11 & 0.12 & 0.13 & 0.16 & 0.27 & 0.39 \\
M-locality    & 0.84 & 0.77 & 0.91 & 0.87 & 0.83 & 0.79 & 0.81 & 0.49 \\
C-generality  & 0.12 & 0.06 & 0.19 & 0.23 & 0.22 & 0.31 & 0.28 & 0.34 \\
C-locality    & 0.07 & 0.11 & 0.06 & 0.09 & 0.44 & 0.47 & 0.50 & 0.29 \\
Temporality   & 0.58 & 0.56 & 0.63 & 0.61 & 0.60 & 0.60 & 0.45 & 0.57 \\
Overall       & 0.21 & 0.16 & 0.23 & 0.25 & 0.41 & 0.39 & 0.53 & 0.30 \\
\bottomrule
\end{tabular}%
}
\caption{Full experiment results on Huatuo 7B. S denotes sequential editing.}
\label{tab:huatuo7b_full_results_singlecol}
\end{table}

\begin{table}[t]
\centering
\scriptsize
\setlength{\tabcolsep}{3.5pt}
\resizebox{\columnwidth}{!}{%
\begin{tabular}{lcccccccc}
\toprule
\textbf{Huatuo 34B} & \textbf{MEND} & \textbf{MEND-S} & \textbf{GRACE} & \textbf{GRACE-S} & \textbf{BE} & \textbf{BE-S} & \textbf{LoRA} & \textbf{LoRA-S} \\
\midrule
Reliability   & 0.27 & 0.31 & 0.61 & 0.53 & 0.96 & 0.64 & 0.95 & 0.94 \\
I-locality    & 0.39 & 0.29 & 0.64 & 0.47 & 0.66 & 0.52 & 0.65 & 0.69 \\
I-generality  & 0.35 & 0.26 & 0.63 & 0.56 & 0.76 & 0.79 & 0.88 & 0.98 \\
T-generality  & 0.51 & 0.24 & 0.19 & 0.15 & 0.57 & 0.59 & 0.67 & 0.77 \\
T-locality    & 0.77 & 0.57 & 0.72 & 0.67 & 0.93 & 0.63 & 0.71 & 0.11 \\
M-generality  & 0.12 & 0.14 & 0.08 & 0.13 & 0.11 & 0.29 & 0.15 & 0.46 \\
M-locality    & 0.93 & 0.87 & 0.95 & 0.88 & 0.97 & 0.77 & 0.79 & 0.57 \\
C-generality  & 0.13 & 0.10 & 0.24 & 0.16 & 0.13 & 0.22 & 0.36 & 0.39 \\
C-locality    & 0.07 & 0.16 & 0.09 & 0.15 & 0.55 & 0.38 & 0.49 & 0.15 \\
Temporality   & 0.57 & 0.56 & 0.62 & 0.59 & 0.57 & 0.47 & 0.52 & 0.63 \\
Overall       & 0.22 & 0.23 & 0.24 & 0.27 & 0.35 & 0.45 & 0.47 & 0.35 \\
\bottomrule
\end{tabular}%
}
\caption{Full experiment results on Huatuo 34B. S denotes sequential editing.}
\label{tab:huatuo34b_full_results_singlecol}
\end{table}
\subsection{Full BELoRA Results (Study on VLM Components and Layers)}

We reported BELoRA results and its sweep on different VLM components and layer depths in analysis. Here we report the full results in Tabs. \ref{tab:component_scope_results} and \ref{tab:component_scope_hmean}.

\begin{table}[t]
\centering
\scriptsize
\setlength{\tabcolsep}{3.5pt}
\resizebox{\columnwidth}{!}{%
\begin{tabular}{llcccccccccc}
\toprule
\textbf{Component} & \textbf{Scope} & \textbf{Task0} & \textbf{Task1} & \textbf{Task1} & \textbf{Task2} & \textbf{Task2} & \textbf{Task3} & \textbf{Task3} & \textbf{Task4} & \textbf{Task4} & \textbf{Task5} \\
 &  & \textbf{relia} & \textbf{loc} & \textbf{gen} & \textbf{gen} & \textbf{loc} & \textbf{gen} & \textbf{loc} & \textbf{gen} & \textbf{loc} &\textbf{temp}  \\
\midrule
LM               & Late & 0.68 & 0.37 & 0.62 & 0.56 & 0.60 & 0.36 & 0.65 & 0.42 & 0.43 & 0.26 \\
                 & Mid  & 0.84 & 0.57 & 0.87 & 0.71 & 0.75 & 0.41 & 0.65 & 0.29 & 0.24 & 0.31 \\
                 & Full & 0.93 & 0.61 & 0.79 & 0.73 & 0.80 & 0.43 & 0.58 & 0.16 & 0.33 & 0.25 \\
\midrule
LM+Proj          & Late & 0.64 & 0.43 & 0.63 & 0.51 & 0.83 & 0.37 & 0.66 & 0.47 & 0.42 & 0.32 \\
                 & Mid  & 0.85 & 0.58 & 0.86 & 0.71 & 0.82 & 0.41 & 0.65 & 0.32 & 0.21 & 0.30 \\
                 & Full & 0.95 & 0.61 & 0.81 & 0.73 & 0.78 & 0.43 & 0.59 & 0.16 & 0.30 & 0.27 \\
\midrule
LM+Proj+Vision   & Late & 0.65 & 0.41 & 0.67 & 0.52 & 0.76 & 0.36 & 0.66 & 0.47 & 0.41 & 0.31 \\
                 & Mid  & 0.87 & 0.58 & 0.88 & 0.71 & 0.82 & 0.41 & 0.66 & 0.33 & 0.21 & 0.29 \\
                 & Full & 0.94 & 0.62 & 0.83 & 0.72 & 0.79 & 0.43 & 0.59 & 0.16 & 0.31 & 0.28 \\
\bottomrule
\end{tabular}%
}
\caption{Results across edited components and layer scope, rounded to two decimals.}
\label{tab:component_scope_results}
\end{table}
\begin{table}[t]
\centering
\scriptsize
\setlength{\tabcolsep}{4pt}
\resizebox{\columnwidth}{!}{%
\begin{tabular}{llccc}
\toprule
\textbf{Component} & \textbf{Scope} & \textbf{HarMean of Loc} & \textbf{HarMean of Gen} & \textbf{HarMean of Everything} \\
\midrule
LM               & Late & 0.48 & 0.47 & 0.45 \\
                 & Mid  & 0.46 & 0.47 & 0.46 \\
                 & Full & 0.52 & 0.36 & 0.42 \\
\midrule
LM+Proj          & Late & 0.54 & 0.48 & 0.49 \\
                 & Mid  & 0.43 & 0.49 & 0.46 \\
                 & Full & 0.51 & 0.36 & 0.42 \\
\midrule
LM+Proj+Vision   & Late & 0.52 & 0.48 & 0.48 \\
                 & Mid  & 0.44 & 0.50 & 0.46 \\
                 & Full & 0.51 & 0.37 & 0.43 \\
\bottomrule
\end{tabular}%
}
\caption{Harmonic mean summary across locality, generality, and all tasks, rounded to two decimals.}
\label{tab:component_scope_hmean}
\end{table}

\subsection{Backbone-Dependent Hyperparameter Sensitivity Results}
We report detailed results for the backbone-dependent hyperparameter sensitivity analysis of memory-based editing in Tab.~\ref{tab:alpha_cross_backbone}.

\begin{table*}[t]
\centering
\scriptsize
\setlength{\tabcolsep}{4pt}
\resizebox{\textwidth}{!}{%
\begin{tabular}{llccc|ccc}
\toprule
\multicolumn{2}{c}{\textbf{}} & \multicolumn{3}{c|}{\textbf{LLaVA-Med}} & \multicolumn{3}{c}{\textbf{Qwen}} \\
\cmidrule(lr){3-5}\cmidrule(lr){6-8}
\textbf{Task} & \textbf{$\alpha$} & \textbf{Trigger Rate} & \textbf{Key Formed Rate} & \textbf{Perf.} & \textbf{Trigger Rate} & \textbf{Key Formed Rate} & \textbf{Perf.} \\
\midrule
\multirow{8}{*}{Task1 Image Locality}
& 0.01 & 0.10 & 1.00 & 0.37 & 0.32 & 0.86 & 0.41 \\
& 0.02 & 0.11 & 1.00 & 0.38 & 0.72 & 0.48 & 0.37 \\
& 0.05 & 0.35 & 0.82 & 0.39 & 0.96 & 0.18 & 0.40 \\
& 0.08 & 0.63 & 0.52 & 0.40 & 0.98 & 0.10 & 0.46 \\
& 0.10 & 0.73 & 0.42 & 0.38 & 0.99 & 0.10 & 0.41 \\
& 0.20 & 0.97 & 0.18 & 0.40 & 1.00 & 0.04 & 0.39 \\
& 0.50 & 1.00 & 0.08 & 0.42 & 1.00 & 0.02 & 0.41 \\
& 0.80 & 1.00 & 0.04 & 0.32 & 1.00 & 0.02 & 0.41 \\
\midrule
\multirow{8}{*}{Task1 Image Generality}
& 0.01 & 0.00 & 0.90 & 0.37 & 0.02 & 0.80 & 0.33 \\
& 0.02 & 0.03 & 0.90 & 0.38 & 0.27 & 0.49 & 0.41 \\
& 0.05 & 0.24 & 0.73 & 0.51 & 0.68 & 0.27 & 0.44 \\
& 0.08 & 0.48 & 0.55 & 0.60 & 0.91 & 0.18 & 0.45 \\
& 0.10 & 0.60 & 0.45 & 0.64 & 0.95 & 0.10 & 0.33 \\
& 0.20 & 0.88 & 0.24 & 0.72 & 1.00 & 0.06 & 0.36 \\
& 0.50 & 1.00 & 0.08 & 0.66 & 1.00 & 0.02 & 0.33 \\
& 0.80 & 1.00 & 0.06 & 0.65 & 1.00 & 0.02 & 0.33 \\
\midrule
\multirow{8}{*}{Task2 Text Generality}
& 0.01 & 0.98 & 0.94 & 0.71 & 0.99 & 0.92 & 0.76 \\
& 0.02 & 1.00 & 0.94 & 0.72 & 1.00 & 0.56 & 0.71 \\
& 0.05 & 1.00 & 0.88 & 0.71 & 1.00 & 0.14 & 0.58 \\
& 0.08 & 1.00 & 0.66 & 0.69 & 1.00 & 0.08 & 0.58 \\
& 0.10 & 1.00 & 0.60 & 0.69 & 1.00 & 0.08 & 0.58 \\
& 0.20 & 1.00 & 0.26 & 0.66 & 1.00 & 0.02 & 0.50 \\
& 0.50 & 1.00 & 0.14 & 0.64 & 1.00 & 0.02 & 0.50 \\
& 0.80 & 1.00 & 0.04 & 0.59 & 1.00 & 0.02 & 0.50 \\
\midrule
\multirow{8}{*}{Task2 Text Locality}
& 0.01 & 0.34 & 0.52 & 0.57 & 0.65 & 0.16 & 0.72 \\
& 0.02 & 0.34 & 0.52 & 0.71 & 0.92 & 0.09 & 0.79 \\
& 0.05 & 0.56 & 0.52 & 0.71 & 1.00 & 0.01 & 0.79 \\
& 0.08 & 0.90 & 0.30 & 0.52 & 1.00 & 0.01 & 0.79 \\
& 0.10 & 0.98 & 0.20 & 0.71 & 1.00 & 0.00 & 0.76 \\
& 0.20 & 1.00 & 0.06 & 0.71 & 1.00 & 0.00 & 0.76 \\
& 0.50 & 1.00 & 0.02 & 0.71 & 1.00 & 0.00 & 0.76 \\
& 0.80 & 1.00 & 0.02 & 0.71 & 1.00 & 0.00 & 0.76 \\
\bottomrule
\end{tabular}%
}
\caption{Example of results from cross-backbone $\alpha$ analysis showing test-time trigger rate, key formed rate (formed keys / total keys), and task performance. Values are rounded to two decimals.}
\label{tab:alpha_cross_backbone}
\end{table*}
\section{Ablation Studies}

\subsection{Additional Analysis on Memory-based Editing Method's Radius ($\alpha$) : Alpha-Induced Memory Collapse and Gradient Incompatibility}
\label{subsec:alpha_memory_collapse}

In the main paper, we observed that increasing the radius in memory-based editors drastically reduces the number of formed keys. This ablation analyzes the consequences of that behavior in more detail.

A larger $\alpha$ makes the editor more permissive, which can increase the chance that a test sample triggers \emph{some} stored edit. However, in sequential editing, this permissiveness may come at a cost: instead of storing edits of different topics in an independent memory slot, the editor may increasingly reuse and train from a small number of existing edited layers. We ask whether larger $\alpha$ improves editing behavior, or whether it instead causes a form of \textbf{memory collapse}, where many edit requests are compressed into too few underlying edited layers.

To study this, we sweep $\alpha \in \{0.01, 0.02, 0.05, 0.08, 0.1, 0.2, 0.5, 0.8\}$ of BE on Huatuo-7B while keeping all other settings fixed. For each run, we log: (1) the number of distinct edited layers used to store the training edits, (2) the effective number of used layers and the dominant-layer share, which quantify many-to-one assignment, (3) test-time trigger rate, and (4) Task0 reliability. To understand whether reused layers group compatible edits or instead force incompatible updates together, we additionally save the training gradient of each edit at the first optimization step and perform post-hoc cosine analysis among edits assigned to the same reused layer.

 In the 200-edit setting, test-time trigger rate is already saturated at 100\% for all $\alpha$, so larger $\alpha$ does not help by rescuing missed edits. Instead, it sharply reduces the number of distinct edited layers, from 131 at $\alpha=0.01$ to 70 at $\alpha=0.02$, 11 at $\alpha=0.05$, and eventually a single layer at $\alpha\geq 0.5$. This collapse is accompanied by a large reliability drop: Task0 exact accuracy falls from 0.93 at $\alpha=0.01$ to roughly 0.47--0.50 for $\alpha\geq 0.05$. Thus, larger $\alpha$ does not improve behavior through better triggering; it mainly compresses many edits into too few parameter slots, reducing effective memory capacity.

Gradient analysis further suggests that this reuse is not benign. If reused layers were grouping only highly compatible edits, same-layer gradients should be consistently aligned. Instead, we find substantially lower-tail incompatibility within reused groups: across reused layers, roughly half of same-layer edit pairs have negative cosine similarity, and more than 85\% fall below 0.1 for larger $\alpha$. This indicates that aggressive reuse forces many weakly compatible or incompatible edits into the same edited layer. Taken together, these findings suggest that $\alpha$ controls a trade-off between permissive routing and memory specificity: beyond a small value, larger $\alpha$ causes \emph{memory collapse}, where many-to-one assignment overwhelms the editor's capacity to preserve distinct corrections. This makes memory-based editing highly sensitive to backbone-specific hyperparameter choices.

\begin{table}[t]
\centering
\scriptsize
\setlength{\tabcolsep}{3pt}
\resizebox{\columnwidth}{!}{%
\begin{tabular}{lcccccc}
\toprule
$\alpha$ & \#Lyrs & Eff.\ Lyrs & Dom.\ Share & Grad-Cos $<0.1$ & Rel. \\
\midrule
0.01 & 131 & 109.56 & 0.06  & 0.692 & 0.930 \\
0.02 & 70  & 27.90  & 0.31  & 0.808 & 0.705 \\
0.05 & 11  & 6.44   & 0.375 & 0.824 & 0.685 \\
0.08 & 6   & 4.13   & 0.375 & 0.850 & 0.625 \\
0.10 & 3   & 2.98   & 0.375 & 0.861 & 0.495 \\
0.20 & 2   & 1.94   & 0.625 & 0.865 & 0.475 \\
0.50 & 1   & 1.00   & 1.00  & 0.867 & 0.465 \\
0.80 & 1   & 1.00   & 1.00  & 0.867 & 0.465 \\
\bottomrule
\end{tabular}%
}
\caption{\textbf{Increasing $\alpha$ induces memory collapse.} As $\alpha$ grows, the editor reuses far fewer distinct edited layers to store the same 200 edits, while trigger rate remains saturated at 100\% for all runs. The resulting many-to-one assignment reduces effective memory capacity and coincides with a sharp drop in Task0 reliability. Gradient lower-tail statistics further show that reused groups are not cleanly coherent: many same-layer edit pairs still have low cosine similarity. ``\#Lyrs'' is the number of distinct edited layers used to store the training edits; ``Eff.\ Lyrs'' is the effective number of used layers computed from the assignment distribution; ``Dom.\ Share'' is the fraction of edits assigned to the most used layer; ``Rel.'' is exact Task0 reliability.}
\label{tab:alpha_memory_collapse}
\end{table}
\subsection{Ablation on Number of Sequential Edits}

In the main paper, we report sequential editing results for every backbone using 200 edits.
Here we report an ablation of number of sequential edits over 1(single editing), 50, 100, 200 on Huatuo-7b only in Tab. \ref{tab:huatuo7b_seq_ablation_mend_grace} and \ref{tab:huatuo7b_seq_ablation_be_lora}.
In the main text, we report results with 200 sequential edits for two reasons. First, this setting serves as a stronger stress test of sequential editing, where interference across edits becomes more pronounced, and method differences are more clearly exposed. Second, we use the same 200-edit setup in the reliability evaluation, and therefore keep this choice throughout the main experiments for consistency and comparability across tasks.
The sequential-edit ablation shows that performance generally degrades as the number of edits increases, confirming that longer edit sequences constitute a harder setting. For MEND and GRACE, the drop is most visible in reliability and locality-related metrics, consistent with increasing interference across edits. BE and LoRA also deteriorate with larger k, but with different failure modes: LoRA is particularly sensitive on several locality metrics at k=200, while BE is relatively more stable on locality yet still loses reliability and generalization as edits accumulate. Notably, k=50 is often much closer to the single-edit regime, whereas k=200 exposes much clearer method differences, supporting our use of 200 sequential edits as a challenging and informative main setting.
\begin{table}[t]
\centering
\scriptsize
\setlength{\tabcolsep}{3.5pt}
\resizebox{\columnwidth}{!}{%
\begin{tabular}{lcccccccc}
\toprule
\textbf{Huatuo 7B} & \textbf{MEND} & \textbf{MEND-200} & \textbf{MEND-50} & \textbf{MEND-100} & \textbf{GRACE} & \textbf{GRACE-200} & \textbf{GRACE-50} & \textbf{GRACE-100} \\
\midrule
Reliability   & 0.18 & 0.13 & 0.21 & 0.19 & 0.73 & 0.54 & 0.67 & 0.62 \\
I-locality    & 0.45 & 0.38 & 0.43 & 0.37 & 0.56 & 0.31 & 0.48 & 0.39 \\
I-generality  & 0.46 & 0.47 & 0.42 & 0.44 & 0.68 & 0.52 & 0.62 & 0.58 \\
T-generality  & 0.38 & 0.19 & 0.27 & 0.33 & 0.24 & 0.21 & 0.23 & 0.23 \\
T-locality    & 0.68 & 0.45 & 0.62 & 0.57 & 0.77 & 0.57 & 0.74 & 0.65 \\
M-generality  & 0.11 & 0.08 & 0.12 & 0.13 & 0.11 & 0.12 & 0.16 & 0.19 \\
M-locality    & 0.84 & 0.77 & 0.84 & 0.79 & 0.91 & 0.87 & 0.95 & 0.86 \\
C-generality  & 0.12 & 0.06 & 0.09 & 0.11 & 0.19 & 0.23 & 0.20 & 0.17 \\
C-locality    & 0.07 & 0.11 & 0.09 & 0.09 & 0.06 & 0.09 & 0.16 & 0.12 \\
Temporal      & 0.58 & 0.56 & 0.61 & 0.56 & 0.63 & 0.61 & 0.63 & 0.65 \\
\bottomrule
\end{tabular}%
}
\caption{Ablation on the number of sequential edits for MEND and GRACE on Huatuo 7B. Plain method names denote single-edit results; suffixes indicate sequential editing with $k$ edits.}
\label{tab:huatuo7b_seq_ablation_mend_grace}
\end{table}
\begin{table}[t]
\centering
\scriptsize
\setlength{\tabcolsep}{3.5pt}
\resizebox{\columnwidth}{!}{%
\begin{tabular}{lcccccccc}
\toprule
\textbf{Huatuo 7B} & \textbf{BE} & \textbf{BE-200} & \textbf{BE-50} & \textbf{BE-100} & \textbf{LoRA} & \textbf{LoRA-200} & \textbf{LoRA-50} & \textbf{LoRA-100} \\
\midrule
Reliability   & 1.00 & 0.67 & 0.76 & 0.69 & 1.00 & 0.57 & 0.86 & 0.68 \\
I-locality    & 0.64 & 0.34 & 0.56 & 0.42 & 0.62 & 0.59 & 0.57 & 0.58 \\
I-generality  & 0.73 & 0.39 & 0.78 & 0.55 & 0.95 & 0.69 & 0.80 & 0.75 \\
T-generality  & 0.52 & 0.48 & 0.68 & 0.60 & 0.73 & 0.62 & 0.70 & 0.65 \\
T-locality    & 0.76 & 0.55 & 0.67 & 0.60 & 0.81 & 0.07 & 0.68 & 0.46 \\
M-generality  & 0.13 & 0.16 & 0.30 & 0.27 & 0.27 & 0.39 & 0.42 & 0.36 \\
M-locality    & 0.83 & 0.79 & 0.80 & 0.78 & 0.81 & 0.49 & 0.67 & 0.55 \\
C-generality  & 0.22 & 0.31 & 0.33 & 0.34 & 0.28 & 0.34 & 0.43 & 0.41 \\
C-locality    & 0.44 & 0.47 & 0.34 & 0.33 & 0.50 & 0.28 & 0.38 & 0.32 \\
Temporal      & 0.60 & 0.60 & 0.63 & 0.55 & 0.45 & 0.57 & 0.50 & 0.48 \\
\bottomrule
\end{tabular}%
}
\caption{Ablation on the number of sequential edits for BE and LoRA on Huatuo 7B. Plain method names denote single-edit results; suffixes indicate sequential editing with $k$ edits.}
\label{tab:huatuo7b_seq_ablation_be_lora}
\end{table}
\section{Algorithm Details and Hyperparameters}
\label{app:algo_details}

This section documents the editing \emph{methods themselves} and the hyperparameters used in our implementation. We focus on the editing mechanism, optimization, and knobs needed for replication. Hyperparameters are selected to obtain strong performance on our benchmark across backbones. Unless otherwise noted, we edit the language model's final-layer MLP. For LoRA, we instead fine-tune all MLP blocks by default, which yields a comparable parameter budget.

\subsection{GRACE}
GRACE (\textbf{G}eneral \textbf{R}etrieval \textbf{A}daptors for \textbf{C}ontinual \textbf{E}diting) is a memory-based editor for lifelong sequential editing. Instead of updating backbone weights, GRACE attaches an adaptor at a chosen layer and stores edits in a discrete codebook. Each entry contains (i) a \emph{key} representation of the edit input at the chosen layer, (ii) a learnable \emph{value} that is applied when the edit is triggered, and (iii) a per-entry \emph{radius} $\epsilon$ that controls when the edit should activate~\cite{hartvigsen2023aginggracelifelongmodel}.

At inference time, GRACE uses the current hidden representation as a query, retrieves the nearest key under a distance function, and applies the associated value only if the query falls within the stored radius; otherwise, it defers to the original model computation. New edits are incorporated by adding a new entry or updating an existing one, and the value is optimized on the edit example while keeping the backbone frozen.

In our runs, we use:
\begin{itemize}
    \item \textbf{Distance metric} \texttt{grace-distance}: \texttt{cosine} (alternative: \texttt{l2}).
    \item \textbf{Initial radius} \texttt{eps-init}: $1.0$.
    \item \textbf{Optimization steps per edit} \texttt{steps}: $100$.
    \item \textbf{Learning rate} \texttt{lr}: $1.0$ (applied to the entry value parameters only).
\end{itemize}
\subsection{MEND}
MEND is a learned editor that predicts parameter updates from gradients. Instead of directly optimizing model parameters for every new edit, MEND trains an \emph{editor network} that takes the gradient signal of an edit loss as input and outputs a parameter update $\Delta\theta$ for a chosen editable parameter subset. The editor is trained across many edits to generalize: at test time, a single forward--backward pass produces $\Delta\theta$ that performs the desired correction while minimizing side effects~\cite{mitchell2022fastmodeleditingscale}.

In each edit, we compute the edit-loss gradient on the editable parameters, feed gradient-derived features into the editor network, and apply the predicted one-step update to obtain edited parameters. To reduce unintended side effects, we additionally evaluate the edited model on a small set of unrelated locality inputs and penalize changes in their predictive distributions.

In our runs, we use:
\begin{itemize}
    \item \textbf{Locality batch size} $|\mathcal{B}_{\mathrm{LOC}}| = 3$ locality inputs per edit.
    \item \textbf{Edit--locality trade-off} $c = 1.0$.
    \item \textbf{Training budget} one pass over the edit-training set.
    \item \textbf{Editor optimizer} Adam with learning rate $\eta$ (we set $\eta=\texttt{0.05}$).
    \item \textbf{Optional stability regularizer} we use a small quadratic penalty on update magnitude with weight $3\times 10^{-4}$.
\end{itemize}

\subsection{BalancEdit (BE)}
 BalancEdit~\cite{guo2025balanceditdynamicallybalancinggeneralitylocality} is a memory-based editor that wraps a selected transformer layer with a codebook of edit entries. Each entry stores (i) a key, computed as the averaged hidden representation of the edit input at the layer preceding the edited layer; (ii) an edited transformation, obtained by fine-tuning the selected layer on the target answer with standard next-token prediction loss; and (iii) an edit-specific activation radius. At inference time, the current hidden representation is compared against stored keys using a distance function, and the edited transformation is activated only when the representation falls within the stored radius; otherwise the original layer is used. Following the paper, the positive anchor for radius estimation is formed by keeping the image fixed and rephrasing the question, while the negative anchor is a black image paired with the original question. The radius is computed as a convex combination of the key-to-positive and key-to-negative distances, controlled by a scalar $\alpha$, which trades off generality and locality. 

 In our runs, we use:
\begin{itemize}
    \item \textbf{Radius coefficient $\alpha$.} We use $\alpha=0.2$ for LLaVA-Med, $0.08$ for Qwen, $0.05$ for Huatuo-7B, and $0.2$ for Huatuo-34B. These values are selected from $\{0.01, 0.02, 0.05, 0.08, 0.1, 0.2, 0.5, 0.8\}$ based on benchmark performance.
    \item \textbf{Learning rate.} We use a learning rate of $0.01$ for all backbones, with gradient clipping at $1.0$ for stability.
\end{itemize}
\subsection{LoRA}
We use \textbf{Lo}w-\textbf{R}ank \textbf{A}daptation (LoRA~\cite{hu2022lora}) of rank 16, alpha 16, dropout 0, AdamW with learning rate 5×$10^{-5}$, batch size 1, gradient clipping 1.0, and 5 epochs by default.
\subsection{BELoRA}
We implement a hybrid memory-based editor that combines BalancEdit-style routing with LoRA-based parameter updates. As in BalancEdit, each stored edit consists of a pooled hidden-state key, an activation radius, and a logical edit identifier. At inference time, the current hidden state is compared against stored keys, and the nearest edit is activated only if its distance falls within the stored radius; otherwise the base module is used. Unlike standard BalancEdit, however, the stored edited parameters are not fully copied to module weights. Instead, for each selected wrapped module, we inject LoRA into matched internal linear layers and store only the learned LoRA parameters for each edit. This yields a multi-block, shared-routing editor in which one routed edit activates coordinated LoRA adapters across all selected blocks.
We use the same $\alpha$ values as in BE and the same optimization hyperparameters as in LoRA.
\section{Computational Costs}
We report the time costs of each editing method per edit on a single NVIDIA A100 GPU (80GB) in Tab.~\ref{tab:time_cost_per_edit}. The unit is seconds (s).
\begin{table}[t]
\centering
\scriptsize
\setlength{\tabcolsep}{3.5pt}
\resizebox{\columnwidth}{!}{%
\begin{tabular}{lcccccccc}
\toprule
& \multicolumn{2}{c}{\textbf{LLaVA-Med}} & \multicolumn{2}{c}{\textbf{Qwen}} & \multicolumn{2}{c}{\textbf{Huatuo 7B}} & \multicolumn{2}{c}{\textbf{Huatuo 34B}} \\
\cmidrule(lr){2-3} \cmidrule(lr){4-5} \cmidrule(lr){6-7} \cmidrule(lr){8-9}
\textbf{Method} & \textbf{Train} & \textbf{Eval} & \textbf{Train} & \textbf{Eval} & \textbf{Train} & \textbf{Eval} & \textbf{Train} & \textbf{Eval} \\
\midrule
MEND  & 2.80  & 2.19  & 1.80  & 2.00  & 2.70  & 1.30  & 3.90  & 1.70  \\
GRACE & 15.80 & 0.45  & 8.90  & 0.40  & 7.60  & 0.36  & 10.20 & 0.33  \\
BE    & 19.88 & 0.78  & 18.56 & 0.82  & 11.88 & 0.37  & 12.71 & 0.56  \\
LoRA  & 6.18  & 0.71  & 1.66  & 0.39  & 2.89  & 0.41  & 3.95  & 0.62  \\
\bottomrule
\end{tabular}%
}
\caption{Average training and evaluation time per edit across methods and backbones (seconds).}
\label{tab:time_cost_per_edit}
\end{table}

\begin{table}[t]
\centering

\label{tab:general_vlm}
\tiny
\setlength{\tabcolsep}{1.1pt}
\renewcommand{\arraystretch}{0.84}
\resizebox{\linewidth}{!}{
\begin{tabular}{l|cccc|cccc|cccc|cccc}
\toprule
& \multicolumn{8}{c|}{Janus Pro-7B} & \multicolumn{8}{c}{Qwen3.5-2B} \\
\cmidrule(lr){2-9}\cmidrule(lr){10-17}
& \multicolumn{4}{c|}{Single} & \multicolumn{4}{c|}{Seq.}
& \multicolumn{4}{c|}{Single} & \multicolumn{4}{c}{Seq.} \\
Task
& M & G & B & L & M & G & B & L
& M & G & B & L & M & G & B & L \\
\midrule
Rel   & 0.55 & 0.63 & \textbf{1.00} & \textbf{1.00} & 0.39 & 0.49 & 0.89 & \textbf{0.97}
      & 0.56 & 0.77 & \textbf{0.99} & \textbf{0.99} & 0.41 & 0.61 & \textbf{0.89} & 0.65 \\
I-loc & 0.38 & 0.46 & \textbf{0.48} & 0.17 & 0.33 & 0.34 & \textbf{0.48} & 0.21
      & 0.62 & \textbf{0.79} & 0.58 & 0.25 & 0.47 & \textbf{0.57} & 0.57 & 0.52 \\
I-gen & 0.72 & 0.50 & 0.78 & \textbf{0.98} & 0.52 & 0.42 & 0.73 & \textbf{0.93}
      & 0.55 & 0.91 & 0.75 & \textbf{0.94} & 0.47 & 0.72 & \textbf{0.75} & 0.74 \\
T-gen & 0.37 & 0.49 & \textbf{0.81} & \textbf{0.81} & 0.29 & 0.43 & \textbf{0.82} & 0.80
      & 0.32 & 0.43 & 0.68 & \textbf{0.81} & 0.26 & 0.31 & 0.51 & \textbf{0.62} \\
T-loc & 0.76 & 0.78 & \textbf{0.96} & 0.78 & 0.70 & 0.73 & \textbf{0.79} & 0.28
      & 0.63 & 0.77 & \textbf{0.92} & 0.57 & 0.58 & 0.73 & \textbf{0.89} & 0.39 \\
M-gen & 0.11 & 0.22 & 0.23 & \textbf{0.37} & 0.08 & 0.18 & 0.22 & \textbf{0.64}
      & 0.17 & 0.28 & 0.25 & \textbf{0.39} & 0.13 & 0.23 & 0.44 & \textbf{0.69} \\
M-loc & 0.89 & 0.78 & \textbf{0.95} & 0.58 & \textbf{0.91} & 0.63 & 0.71 & 0.14
      & 0.88 & 0.88 & \textbf{0.98} & 0.77 & \textbf{0.90} & 0.86 & 0.83 & 0.44 \\
C-gen & 0.01 & 0.12 & 0.58 & \textbf{0.74} & 0.02 & 0.10 & 0.60 & \textbf{0.83}
      & 0.27 & 0.33 & 0.42 & \textbf{0.88} & 0.23 & 0.24 & 0.39 & \textbf{0.51} \\
C-loc & 0.26 & 0.15 & \textbf{0.58} & 0.21 & 0.34 & 0.33 & \textbf{0.63} & 0.23
      & 0.16 & 0.27 & \textbf{0.41} & 0.34 & 0.37 & 0.15 & \textbf{0.46} & 0.16 \\
Temp  & \textbf{0.42} & 0.38 & 0.28 & 0.29 & \textbf{0.51} & 0.44 & 0.24 & 0.47
      & 0.30 & 0.21 & \textbf{0.37} & 0.32 & \textbf{0.39} & 0.26 & 0.30 & 0.24 \\
\midrule
Overall & 0.08 & 0.31 & \textbf{0.52} & 0.41 & 0.12 & 0.30 & \textbf{0.49} & 0.36
        & 0.33 & 0.43 & \textbf{0.52} & 0.50 & 0.33 & 0.34 & \textbf{0.53} & 0.40 \\
\bottomrule
\end{tabular}
}
\caption{ Single and sequential editing performance. M: MEND; G: GRACE; B: BalancEdit; L: LoRA.}
\end{table}

\begin{table}[t]
\centering
\label{tab:cone_pairs}
\scriptsize
\setlength{\tabcolsep}{3pt}
\renewcommand{\arraystretch}{0.88}
\begin{tabular}{l|ccc|ccc}
\toprule
& \multicolumn{3}{c|}{Gemma family} & \multicolumn{3}{c}{LLaVA family} \\
\cmidrule(lr){2-4}\cmidrule(lr){5-7}
Dataset & Gemma & MedGemma & $\Delta$ & LLaVA & LLaVA-Med & $\Delta$ \\
\midrule
VQA-RAD & .920 & .949 & \textbf{+.029} & .842 & .932 & \textbf{+.090} \\
SLAKE   & .817 & .872 & \textbf{+.055} & .825 & .888 & \textbf{+.063} \\
Mix     & .839 & .890 & \textbf{+.051} & .827 & .892 & \textbf{+.065} \\
\bottomrule
\end{tabular}

\caption{Mean resultant length $R$ (cone tightness).}
\end{table}

\subsection{Additional Results on Non-Medical VLM and Native Multimodal VLM}
We additionally \emph{evaluate all four editors on Qwen3.5-2B and Janus Pro-7B.}
</invoke> Results in~\cref{tab:general_vlm} confirm the gradient-vs-memory trade-off is a general characteristics of multimodal editing rather than medical-VLM-specific. In sequential editing, LoRA exhibits its signature T-locality and M-locality collapse, while BalancEdit remains the highest overall performer on both VLMs. Editing on general VLMs also appears easier for some editors: BE overall is higher on general models (0.49–0.53) than on medical VLMs (0.39–0.46); LoRA's T-locality collapse is less severe on general (0.28–0.39) than on medical models(0.03–0.11). 
This may be explained by cone effect analysis in~\cref{tab:cone_pairs}, where we find medical tuning worsens the cone effect from general VLMs, potentially indicating less anisotropic representations of general VLMs admit better localized editing.
\subsection{More On Cone Effect}We conducted controlled base-vs-medical pair comparisons on two VLM families. The cone effect is inherited from the base VLM and further amplified by medical fine-tuning: medical alignment consistently increases anisotropy , with the effect more pronounced in LLaVA and present but smaller in Gemma. This directly explains why our evaluation shows easier editing on general VLMs, and why editor hyperparams of VLMs do not transfer between backbones without re-tuning.
\end{document}